\documentclass[review]{elsarticle}
\usepackage{lineno,hyperref, bm, float, graphicx, subcaption, amssymb, amsmath, multirow}
\modulolinenumbers[5]
\usepackage{xcolor}
\usepackage{algorithm}
\usepackage{algpseudocode}

\journal{Journal of Renewable Energy Templates}

\begin{document}

\begin{frontmatter}
\title{Reinforcement learning-enhanced genetic algorithm for wind farm layout optimization}
%风电场布局优化中的强化学习驱动基因算法
% \author[mymainaddress,mysecondaryaddress]{Guodan Dong\fnref{myfootnote1}}
% \fntext[myfootnote1]{These two authors contributed equally.}

% \author[mymainaddress,mysecondaryaddress]{Zhaobin Li\fnref{myfootnote2}}
% \fntext[myfootnote2]{These two authors contributed equally.}
\author[mymainaddress]{Guodan Dong}
\author[mymainaddress,qinaddress]{Jianhua Qin}
% \ead{qjhnjust@qq.com}
\author[yangaddress,yang_secondaryaddress]{Chutian Wu}
% \author[shenaddress,mymainaddress]{Wenzong Shen}
\author[mymainaddress]{Chang Xu}
% \author[mymainaddress]{Chang Xu\corref{mycorrespondingauthor}}
% \ead{zhuifengxu@hhu.edu.cn}

\author[yangaddress,yang_secondaryaddress]{Xiaolei Yang\corref{mycorrespondingauthor}}
\cortext[mycorrespondingauthor]{Corresponding author}
\ead{xyang@imech.ac.cn}

\address[mymainaddress]{College of Renewable Energy, Hohai University, Changzhou, 213200, China}

\address[qinaddress]{Nanjing University of Science and Technology, Jiangyin, 214443, China}
% \address[shenaddress]{College of Electrical, Energy and Power Engineering, Yangzhou University, Yangzhou 225127, China}

\address[yangaddress]{The State Key Laboratory of Nonlinear Mechanics, Institute of Mechanics, Chinese Academy of Sciences, Beijing 100190, China}
\address[yang_secondaryaddress]{School of Engineering Sciences, University of Chinese Academy of Sciences, Beijing 100049, China}

\begin{abstract}
    A reinforcement learning-enhanced genetic algorithm (RLGA) is proposed for wind farm layout optimization (WFLO) problems. While genetic algorithms (GAs) are among the most effective and accessible methods for WFLO, their performance and convergence are highly sensitive to parameter selections. To address the issue, reinforcement learning (RL) is introduced to dynamically select optimal parameters throughout the GA process. To illustrate the accuracy and efficiency of the proposed RLGA, we evaluate the WFLO problem for four layouts (aligned, staggered, sunflower, and unstructured) under unidirectional uniform wind, comparing the results with those from the GA. RLGA achieves similar results to GA for aligned and staggered layouts and outperforms GA for sunflower and unstructured layouts, demonstrating its efficiency. The sunflower and unstructured layouts' complexity highlights RLGA's robustness and efficiency in tackling complex problems. To further validate its capabilities, we investigate larger wind farms with varying turbine placements ($\Delta x = \Delta y = 5D$ and $2D$, where $D$ is the wind turbine diameter) under three wind conditions: unidirectional, omnidirectional, and non-uniform, presenting greater challenges than the test case. The proposed RLGA is about three times more efficient than GA, especially for complex problems. This improvement stems from RL's ability to adjust parameters, avoiding local optima and accelerating convergence. 
\end{abstract}

\begin{keyword}
\texttt{} Reinforcement learning \sep Genetic algorithm \sep Wind farm layout optimization
\end{keyword}
\end{frontmatter}
% \linenumbers
\section{Introduction}
%%下面是根据文献写的，已改
Modern wind farms, composed of tens to hundreds of multi-megawatt turbines, collectively supply low-cost energy to worldwide. Over their decades-long operation, these turbines interact with turbulent atmospheric flows, and are influenced by the wakes of upstream turbines~\cite{veers2019grand, meyers2022wind}. Wakes can decrease power output and increase fatigue load for the downstream turbine~\cite{yang2019review}, increasing the need for comprehensive investigation into wind farm layout optimization (WFLO), which aims to strategically position wind turbines to maximize energy production while minimizing the wake effects~\cite{gonzalez2014review}. 
Because of the large number of variables, the optimization efficiency is one of the challenges for WFLO problems. 
%
%sThe optimization challenge involves several critical components, including a turbine wake model, an optimization algorithm, and objective functions. 
%The primary challenge in WFLO lies in 
Properly modeling the velocity deficit induced by turbine wakes is critical for WFLO. 
%, which result in reduced wind speeds and increased turbulence. To simulate turbine wakes, 
Various wake models of differing fidelities have been employed, from low-fidelity analytical models to high-fidelity computational fluid dynamics (CFD) approaches. These CFD methods include simplified Navier-Stokes models~\cite{larsen2007dynamic}, Reynolds-Averaged Navier-Stokes (RANS) models~\cite{iungo2015data}, and large-eddy simulations (LES)~\cite{li2022onset}. However, CFD simulations require thousands of CPU hours for a single run, making analytical wake models the preferred choice in WFLO. 
In analytic wake models, the Jensen ``top-hat'' wake model was initially proposed based on the mass conservation equation~\cite{jensen1983note, katic1987simple}. In recent years, more accurate Gaussian wake models have been developed~\cite{bastankhah2014new, gao2016optimization}. To further enhance the precision and adaptability of Gaussian-based models, additional real-world wake characteristics have been incorporated, such as the double-Gaussian velocity profile observed in the near wake~\cite{keane2021advancement}, three-dimensional flow effects~\cite{sun2018study}, yaw misalignment of wind turbines~\cite{bastankhah2016experimental}, and the Coriolis effect~\cite{cheng2019new}.
Among these, the Jensen wake model~\cite{jensen1983note}, is widely utilized in WFLO problems~\cite{mosetti1994optimization, wu2021design, yu2024teaching} and will be also employed in this work.

%% Physics-informed layouts
Many studies have employed optimization algorithms based on discrete variables, often defining potential turbine locations on a Cartesian grid~\cite{rajper2012optimization, parada2017wind}. Building on the pioneering work of Mosetti et al.~\cite{mosetti1994optimization} and our recent work~\cite{wu2021design}, this study also investigate the WFLO based on discrete variables.
The complexity of WFLO is immense; even with fewer than 30 turbines, the solution space during the optimization can exceed $10^{44}$. Additionally, it is a mixed-integer, highly non-convex problem characterized by numerous local optima~\cite{herbert2014review}.
To tackle these complex and nonlinear optimization problems, researchers have increasingly turned to meta-heuristic (MH) algorithms inspired by natural evolution and swarm intelligence~\cite{wilson2018evolutionary}. MHs can be broadly classified into three categories: evolution-based algorithms, swarm intelligence algorithms, and others. Evolution-based algorithms minic the biological processe, such as heredity, variation, and selection to explore the solution space in search of an optimal configuration. Examples include the genetic algorithm (GA)~\cite{barrera2012optimal} and dfferential evolution (DE)~\cite{gao2017understanding}. Swarm intelligence algorithms, on the other hand, mimic the collective behavior observed in nature, such as particle swarm optimization (PSO)~\cite{lei2022adaptive}, and ant colony optimization (ACO)~\cite{gao2016ant}. 
%
%Numerical studies~\cite{yang2014investigation} and experimental research~\cite{chamorro2011turbulent, hamilton2015wind} have demonstrated that a staggered turbine array can generate more power than an aligned turbine array with the same turbine density. Similar to the recent WFLO work in Ref.~\cite{wu2021design} four types of physics-informed turbine layout configurations are considered: (a) aligned, (b) staggered, (c) sunflower, and (d) unstructured. 
%
%Other MHs, such as simulated annealing (SA)~\cite{rivas2009solving}, tabu search (TS)~\cite{glover1986future}, and artificial immune algorithm (AIA)~\cite{holland1992adaptation}, are based on a range of heuristic principles.
%
In recent years, novel algorithms have been introduced to address the WFLO challenges. For example, Li et al.~\cite{li2022discrete} proposed a discrete complex-valued pathfinder algorithm (DCPFA) utilizing a diploid encoding scheme for individual representation. Lei et al.~\cite{lei2022adaptive} developed an adaptive genetic learning particle swarm optimization (AGPSO) algorithm that dynamically adjusts the position of underperforming wind turbines to enhance wind farm efficiency.

The performance of the GA-like algorithms is highly sensitive to parameter settings, such as crossover probability, mutation probability, and number of population per solution. Proper parameter tuning is crucial for optimizing algorithm performance, given their sensitivity to these settings~\cite{elkinton2007optimization}. Different problems require specific adjustments, often necessitating numerous trials and optimizations. To overcome these drawbacks, reinforcement learning (RL) techniques can be integrated to optimize parameters during the MH process. %By combining RL with search algorithms, researchers can capitalize on the strengths of both approaches, improving search efficiency and overall performance~\cite{hsieh2016aq}. 
Recently, machine learning techniques have increasingly been integrated into evolutionary algorithms (EAs) to enhance their performance in optimization tasks. These techniques include supervised, unsupervised, semi-supervised, and RL, which enable agents to learn and select optimal actions within specific contexts to maximize rewards. For instance, Monte Carlo tree search have been enhanced with RL to improve search efficiency and solution quality~\cite{bai2022wind}. Multi-agent RL frameworks, like the actor-critic approach, have been employed to increase wind farm power output by dividing large farms into smaller groups~\cite{dong2023reinforcement}. RL has also been integrated with other EAs, such as Thompson sampling, to minimize power plant losses~\cite{vyshnav2022reinforcement}.
Deep RL models have been applied to WFLO, defining actions, states, and rewards, and using deep convolutional neural networks to optimize these elements~\cite{li2022deep}. Novel deep RL-based approaches offer flexible and efficient solutions for optimizing wind farm power generation~\cite{dong2021intelligent}. Furthermore, decentralized RL frameworks, where each turbine makes independent yaw decisions, have been used to maximize output~\cite{deng2023decentralized}. Advanced techniques like composite experience replay in RL have been employed to enhance power production~\cite{dong2021composite}. Additionally, RL-based methods have been developed for yaw-based wake steering, adapting to changing wind conditions to maximize energy capture~\cite{stanfel2020distributed}. Knowledge-assisted RL frameworks have also been introduced to ensure maximum power output through various knowledge-assisted methods~\cite{zhao2020cooperative}. Finally, multiplayer deep RL approaches have been proposed as data-driven control schemes for optimizing wind farm power output~\cite{dong2022data}. % motivating the development of more competitive algorithms.

%Considering the tranmendous function of RL, this work aims to propose a RL-driven GA (RLGA) method, which allows researchers to invest more effort in algorithm and strategy design. The RL approach helps the GA find higher-quality solutions through intelligent decision-making. Such an approach has strong application prospects and can effectively deal with various types of complex situations. For the method, this is the first time that RL is embed into GA for the WFLO problem, and this research idea can also be applied to other combinatorial optimization problems.

The advancements outlined above demonstrate the significant potential of RL in addressing the WFLO problem. While GA are among the most popular and easy-to-use optimization techniques~\cite{mosetti1994optimization, grady2005placement, parada2017wind, wu2021design}, they are highly sensitive to parameter selection, which can lead to local minima. To mitigate this issue, researchers often resort to using large populations and generations. However, this approach can become time-consuming when dealing with optimization problems that have extensive parameter spaces, such as WFLO for large wind farms.
his work proposes an RL-enhanced genetic algorithm (RLGA) method that uses a significantly smaller population size per generation to achieve comparable or improved results, enhancing the GA's convergence efficiency and facilitating escape from local minima through dynamic selection of GA parameters.
%To accelerate the convergence of GA and escape local minima, this work proposes a RL-enhanced GA (RLGA) method that utilizes a very small number of populations in each generation. 
The RL action spaces consist of two choices for parent mating, four options for crossover types, and four alternatives for mutation rates. During each GA iteration, these RL actions are selected based on previous generations, and the Q-table is updated using the Bellman equation. This RL approach enables the GA to find higher-quality solutions through intelligent decision-making.
This method not only shows strong application prospects but also effectively addresses various complex situations. To the best of our knowledge, this is the first work to integrate RL into GA for the WFLO problem, and the proposed method can be applied to other combinatorial optimization problems.
%It is the first instance of integrating RL into GA for the WFLO problem, and this research idea can be extended to other combinatorial optimization challenges.

The paper is organized as follows: Section \ref{sec:methods} describes the methods and setups, including the Jensen wake model (Section \ref{sec:Jensen_model}), the optimization algorithm (Section \ref{sec:opt_methods}), and the setups (Section \ref{sec:Potential_layout}). Results are presented and discussed in Section \ref{sec:results}, and the main findings are summarized in Section \ref{sec:conclusion}.

\section{Methods and setups} \label{sec:methods}
This section provides a comprehensive illustration of the methodologies employed, including the wake modelling (the Jensen wake model, the wake superposition method, the calculation of relative distance in align with the wind direction), the optimization methods (GA, RL, and RLGA), and the setups (three wind conditions, three wind farm sets, and four physics-informed potential layouts).

\subsection{Wake modeling} \label{sec:Jensen_model}
\begin{figure}[H]
    \centering
    \includegraphics[width=0.42\textwidth]{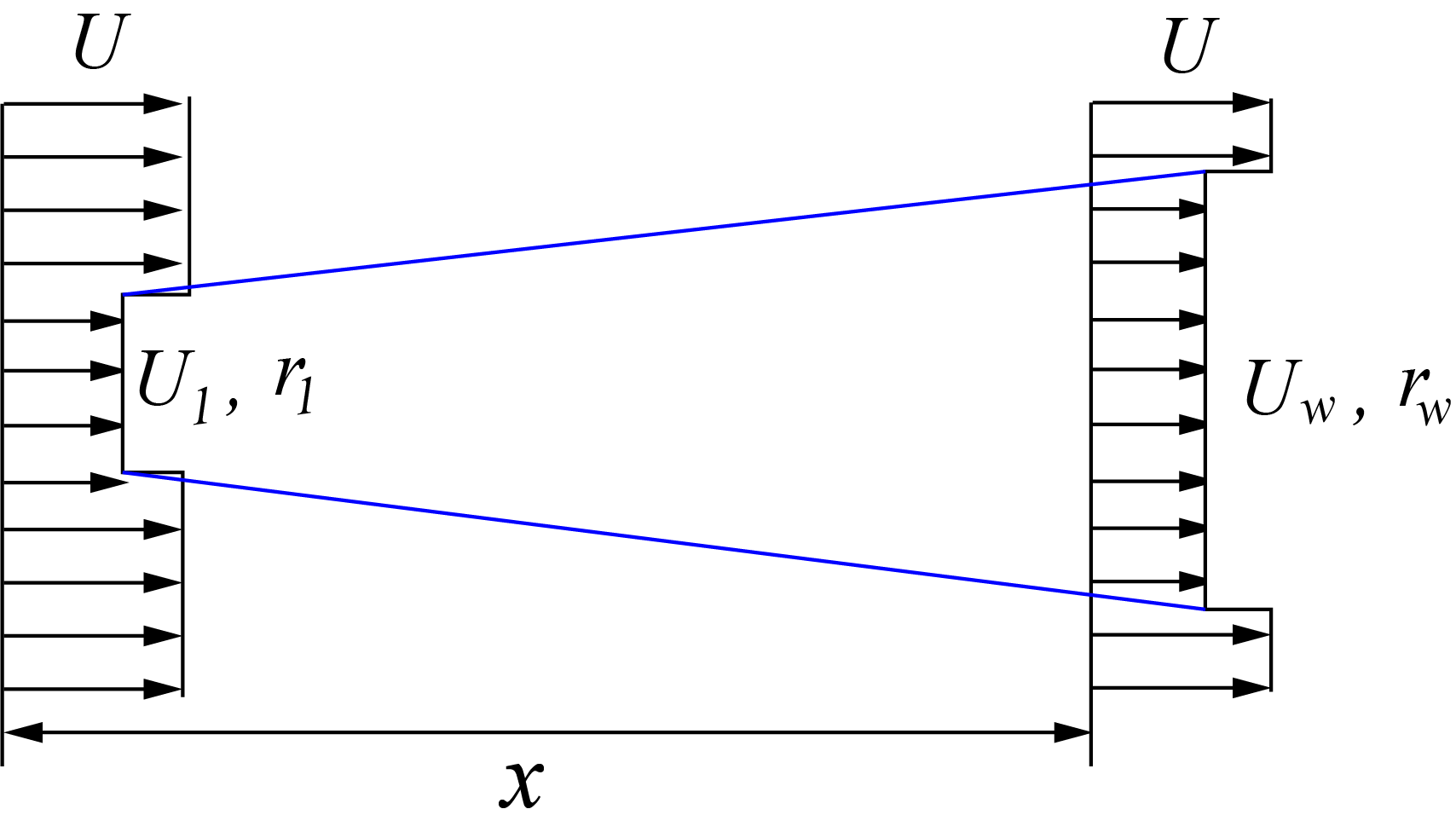}
    \includegraphics[width=0.47\textwidth]{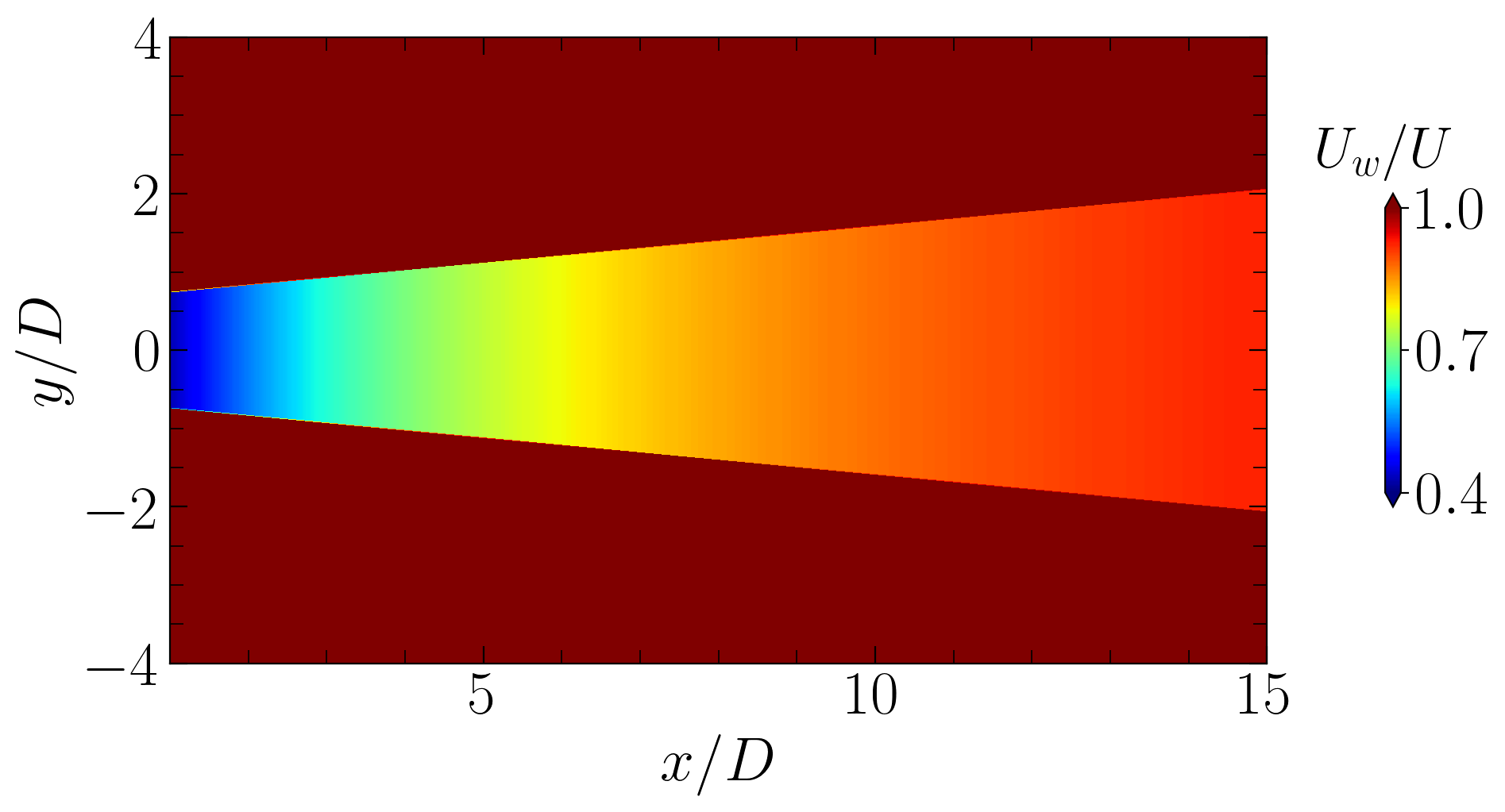}
    \caption{The schematic (left) and contour (right) of the Jensen single wake model.}
    \label{fig:Jensen_model}
\end{figure}

In this work, the Jensen wake model~\cite{jensen1983note} is used, which assumes a uniform velocity profile behind the rotor with linear wake expansion due to energy entrainment.
The schematic and velocity contours of the Jensen single wake model is shown in Fig. \ref{fig:Jensen_model}.
Here, $U$ is the free-stream wind velocity, $U_1$ denotes the wake wind speed immediately behind the turbine, $r_1$ is the initial wake radius, and $r_w$ and $U_w$ represent the wake radius and wake velocity, respectively, at a downwind distance $x$. According to the law of momentum conservation, the wake radius ($r_w$) expands linearly with downwind distance ($x$)~\cite{schepers2003endow}, such that $r_w = r_1 + \alpha_e x$. Here, $\alpha_e$ is the entrainment constant, defined as:
\begin{equation}
\alpha_e = \frac{0.5}{\text{ln}(z_h / z_0)},    
\end{equation}
where $z_h$ is the hub height and $z_0$ is the surface roughness length.

According to the deviation of the Jensen wake model shown in the appendix, the Jensen single wake model can be written as:
\begin{equation}
    \frac{\Delta U}{U} = \frac{1-\sqrt{1-C_T}}{(1+\alpha_e X/r_1)^2},
    \label{eq:jensen_model}
\end{equation}
where $\Delta U = U - U_w$. Note that the Jensen model employed in this work is the same as that used by Frandsen~\cite{frandsen1992wind}, differing from the original Jensen wake model~\cite{jensen1983note, katic1987simple} in the calculation of the entrainment constant $\alpha$ and the initial wake radius $r_1$. 

To account for the impact of wakes from upwind turbines on velocity deficit, the quadratic sum method~\cite{goccmen2016wind}, which assumes the conservation of the mean kinetic energy deficit during wake interaction, is employed. Thus, the wind speed $U_i$ at turbine $i$ is determined as follows:
\begin{equation}
    U_i = U - \sqrt{\sum_j^{n}(U -  U_{ij})^2 },
\end{equation}
where $U_{ij}$ represents the wind speed at turbine $i$ influenced by the wake from turbine $j$, with the summation encompassing the $n$ turbines located upwind of turbine $i$. Various wake superposition methods can be applied, and comprehensive reviews of these methods are available in Refs.~\cite{stevens2017flow, porte2020wind}.

Besides, when the downstream turbine is only partially within the wake, the wake deficit should be scaled according to the fraction of the rotor area~\cite{feng2014wind},denoted as $A_w$, that lies within the upstream turbine's wake~\cite{feng2014wind}.
Then, the effect of the corresponding deficit must be reduced as:
\begin{equation}
    U_i = U - \sqrt{\sum_j^{n} \left[ \frac{A_w}{\pi r^2}(U -  U_{ij})\right]^2 },
\end{equation}
\begin{figure}[H]
    \centering
    \includegraphics[width=0.4\textwidth]{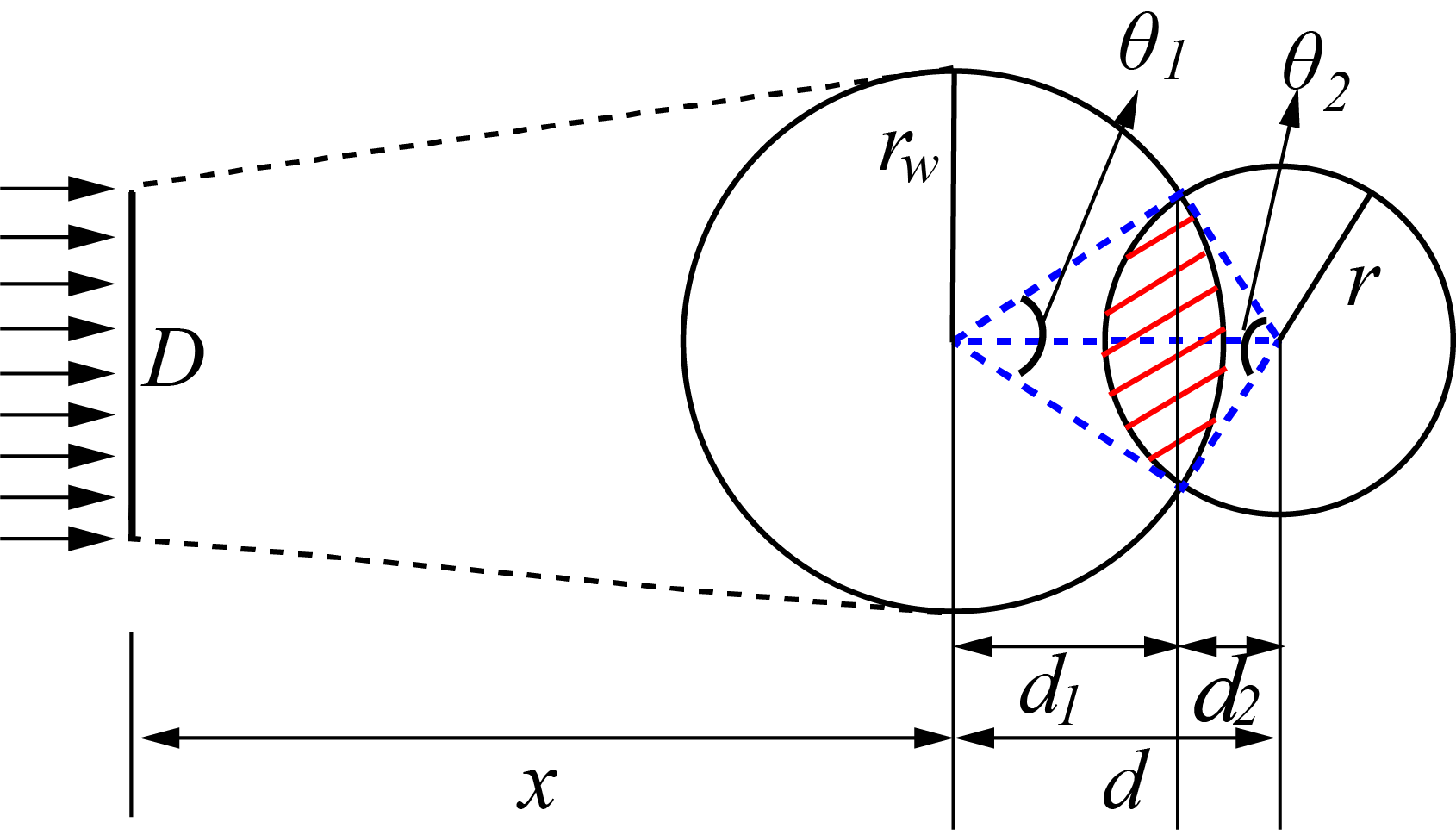}
    \caption{Overlapped area between a wind turbine rotor and a wake stream.}
    \label{fig:Overlap}
\end{figure}

Fig.~\ref{fig:Overlap} illustrates the schematic of wake interference between an upwind turbine $j$, and a downwind turbine $i$, which is similar to the previous studies~\cite{abdulrahman2017investigating, gonzalez2010optimization}. In the figure, $r_{w}$ represents the radius of the wake from the $j^{th}$ turbine as it reaches the $i^{th}$ turbine, and $A_{w}$ denotes the overlapped area between the $j^{th}$ turbine's wake and the rotor of the downwind turbine. Then $A_w$ can be calculated as:
\begin{equation}
    A_w = \left\{
\begin{array}{lcl}
0 &   & {d > r + r_w} \\
\pi r^2 &   & {d <= |r_w - r|} \\
\frac{1}{2} r_w^2 (\theta_1 - \text{sin}\theta_1) + \frac{1}{2} r^2 (\theta_2 - \text{sin}\theta_2) &   & {d >  |r_w - r|}
\end{array} \right.
\end{equation}
Here $d_1 = \frac{r_w^2 + d^2 - r^2}{2d}, \, d_2 = d - d_1, \, \theta_1 = 2 \text{cos}^{-1}\frac{d_1}{r_w}, \, \theta_2 = 2 \text{cos}^{-1}\frac{d_2}{r}$.

In the present WFLO problem, three different wind conditions blowing from different directions are considered. To accurately calculate the velocity deficit factor for turbine $i$ under the influence of multiple wakes, it is essential to have precise information regarding the relative distances between turbines aligns with the wind direction $\theta$.
Consider the scenario in which turbine $i$ is located within the wake of turbine $j$ when the wind blows in the direction of $\theta$. The coordinates of turbine $i$ in a northwest coordinate system, with the origin at turbine $j$, are given as $(x_i, y_i)$. Through a passive transformation rotation matrix, one can determine the coordinates of turbine $i$ relative to a coordinate system where the x-axis aligns with the wind direction $\theta$ and the origin remains at turbine $j$.  Consequently, the coordinates $(x_i^{'}, y_j^{'})$  of turbine $i$ relative to the wind direction $\theta$ can be derived using the following equation:
\begin{equation}
(x'_i, y'_i) = (x_i, y_i) \begin{bmatrix} \cos(\theta) & \sin(\theta) \\ -\sin(\theta) & \cos(\theta) \end{bmatrix}.
\end{equation}

% For instance, the relative radial distances between points A and B (shown in Fig.~\ref{fig:Rotate}) can be calculated as follows:
% \begin{align}
%     \Delta y = dx \, \text{sin}(\theta) + dy \, \text{cos}(\theta), \\
%     \Delta x = dx \, \text{cos}(\theta) - dy \, \text{sin}(\theta).
% \end{align}

% \begin{figure}[H]
%     \centering
%     \includegraphics[width=0.35\textwidth]{Rotate.png}
%     \caption{Schematic of calculating the axial distances.}
%     \label{fig:Rotate}
% \end{figure}

After determining the relative distances between all turbines within the wind farm, the next step involves calculating the velocity deficit for each turbine. Following this, the power output of all turbines is computed based on the theoretical wind turbine power curve, as described in Refs.~\cite{mosetti1994optimization, grady2005placement, parada2017wind}. %Alternatively, more sophisticated methods for power calculation may be employed, such as using the wind turbine experimental power curve [6,28] or those incorporated in predictive control strategies [4,35,36]. 
Once the wind farm's total power output is determined, it is adjusted by the probability of different wind speeds and directions. Finally, the cost of energy is calculated based on this weighted power output.

The turbine parameters used in the present work is the same as the previous studies in Refs.~\cite{mosetti1994optimization, grady2005placement, wu2021design}, with $z_{hub} = 60 \,\text{m}, \, D =  40 \,\text{m},\, z_0 = 0.3\,\text{m, and } C_T = 0.88$. Then the axial induction factor $a = 0.33$ can be obtained according to Eq.~(\ref{eq: a}). The power  is defined by the equation: $P = 0.3 U^3 \, \text{(kw)}$, enabling the calculation of the total power extracted by the wind farm as:
\begin{equation}
    P_{total} = \sum_i^n 0.3 U_i^3,
\end{equation}
where $N$ denotes the total number of turbines.

\subsection{Optimization method} \label{sec:opt_methods}
% When designing a wind farm, the optimal number of turbines is often unknown or difficult to estimate. Fixing the number of turbines based on the rated power capacity of the wind farm may lead to either an overestimation or underestimation of the site's full wind energy potential. Conversely, treating the number of turbines as a variable to be optimized in WFLO problems allows for the possibility of achieving an optimal layout that maximizes the site's energy potential while minimizing the adverse effects of turbine wakes.

Various objective functions have been utilized in the literature \cite{elkinton2007optimization} for WFLO problems. In this study, we adopt the objective function outlined in Refs. \cite{mosetti1994optimization, grady2005placement, wu2021design}, defined as the cost per unit of power produced, which is expressed as:
\begin{equation}
    f_\text{obj} = \frac{P_{\text{total}}}{\text{cost}},
\end{equation}
where the total annual cost is provided in Ref.~\cite{wu2021design}:
\begin{equation}
    \text{cost} = N \left( \frac{2}{3} + \frac{1}{3} \exp\left(-0.00174 N^2\right) \right)
\end{equation}
This cost function is non-dimensionalized using the annual cost of a single turbine and assumes a maximum cost reduction of 1/3 for each turbine~\cite{mosetti1994optimization}. And the fitness ($F$) in defined as the reciprocal of the objective function as:
\begin{equation}
    F = \frac{1}{ f_\text{obj} -   f_\text{obj, ideal}},
\label{eq:fitness}
\end{equation}
where, $f_\text{obj, ideal}$ is the ideal objective fitness, which will be specified in the following section.

\subsubsection{Genetic algorithm}
A genetic algorithm (GA) is an optimization method inspired by natural selection, where a population of candidate solutions evolves through selection, mating, crossover, and mutation. Each candidate is evaluated by a fitness function, and better solutions are selected as parents. Crossover combines genes from parents to explore new solutions, while mutation introduces random changes to maintain diversity and prevent premature convergence.

\begin{algorithm}
\caption{Genetic Algorithm}\label{alg:genetic_algo}
\small  % Adjusts the font size of the algorithm
\setlength{\abovecaptionskip}{0pt}  % Reduces space above the caption
\begin{algorithmic}[1]
    \State \textbf{Input:} Population size $N_{\text{p}}$, number of generations $N_g$, number of parents mating $P_C$, mutation rate $M_C$, crossover rate $C_C$
    \State \textbf{Output:} Best individual $x_{\text{best}}$    
    \For{$n_g = 1$ to $N_g$}
        \State \textbf{Selection:} Select $N_{\text{p}}$ individuals from the population based on fitness
        \State \textbf{Mating:} With $P_C$, perform parents mating
        \State \textbf{Crossover:} With $C_C$, perform crossover on individuals to create offspring
        \State \textbf{Mutation:} With probability $M_C$, mutate the offspring
        \State Evaluate the fitness of the new offspring
        \State Replace the old population with the new offspring population
    \EndFor
    \State \textbf{Return:} Best individual $x_{\text{best}}$ from the final population
\end{algorithmic}
\end{algorithm}

GA has been extensively used in WFLO~\cite{mosetti1994optimization, grady2005placement, parada2017wind, wu2021design}. In general, GA operate on encoded representations of parameters rather than the parameters themselves. In binary-coded GAs, individuals are represented as binary strings composed of ones and zeros.
In the present WFLO problems, given the characteristics of a potential wind farm site—such as area, wind conditions, and other constraints—potential turbine positions are designed. Using these potential positions, a population of different turbine layouts is generated, represented as binary arrays where 0 indicates no turbine and 1 indicates the presence of a turbine. The power outputs for each layout are then calculated using the Jensen wake model as shown in section~\ref{sec:Jensen_model}. 
Based on the fitness values, parents are selected to create a new generation of turbine positions through genetic manipulation. At the end of the optimization, the turbine layout with the highest fitness is obtained. Importantly, the total number of turbines, which is not predefined in the optimization, is also determined by the procedure. This approach is valuable as the optimal number of turbines is often unknown during the initial stages of wind farm design.
The process of performing the GA is outlined in Algorithm~\ref{alg:genetic_algo}, with the corresponding flowchart depicted on the Fig.~\ref{fig:flow_chart} .

\subsubsection{Reinforcement learning}
\begin{figure}[H]
    \centering
    \includegraphics[width=0.45\textwidth]{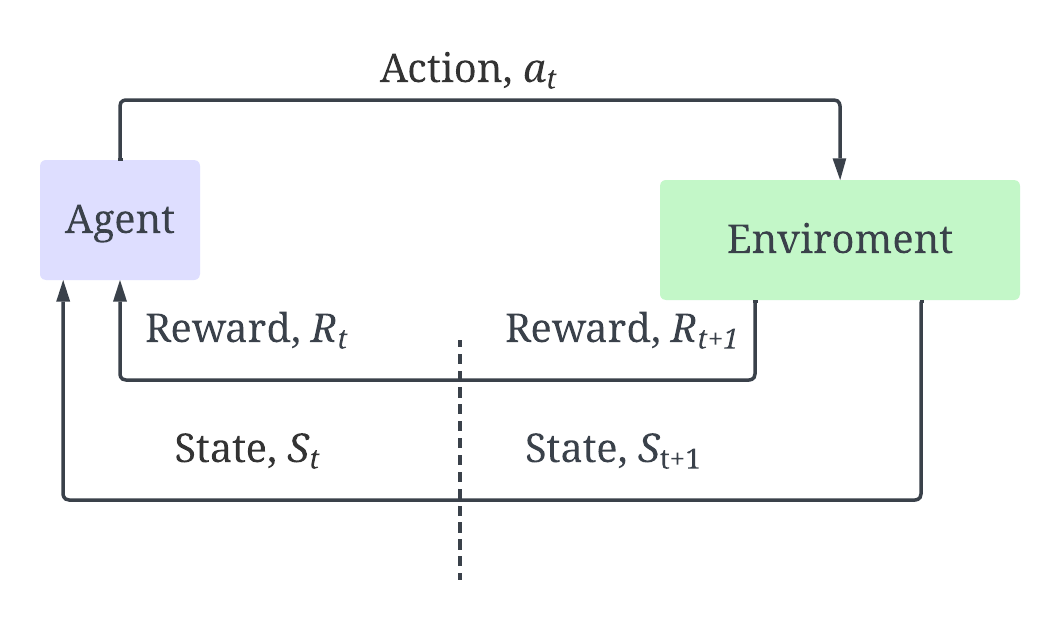}
    \caption{Block diagram schematic for RL.}
    \label{fig:RL_diagram}
\end{figure}

Reinforcement learning (RL) is a decision-making framework that enables an agent to learn optimal strategies through interaction with an environment, as shown in Fig.~\ref{fig:RL_diagram}. Among RL methods, Q-learning is notable for its simplicity and effectiveness, widely used in models like Deep Q-Networks (DQN) and Deep Reinforcement Learning (DRL) for adaptive decision-making based on Q-values, which represent the expected rewards for actions in different states~\cite{ kumar2020conservative}. Q-learning is often characterized by the quintuple $\langle A, S, R, L, E \rangle$, representing actions, states, rewards, the learning algorithm, and the environment~\cite{watkins1992q}.
\begin{algorithm}[H]
\caption{Q-learning Algorithm}\label{alg:q_learning}
\small  % Adjusts the font size of the algorithm
\setlength{\abovecaptionskip}{0pt}  % Reduces space above the caption
\begin{algorithmic}[1]
    \State Initialize Q-table $Q(S, A)$ with small random values
    \State Set parameters: learning rate $\alpha$, discount factor $\gamma$, exploration rate $\epsilon$
    \State Initialize state $S$
    \For{each episode}
        \State Reset environment to initial state $S$
        \While{not end of episode}
            \State Choose action $A$ using policy derived from $Q$ (e.g., $\epsilon$-greedy)
            \State Take action $A$, observe reward $R$ and new state $S'$
            \State Update Q-table according to Eq.~(\ref{eq:Bellman})
            \State Update state $S$ to $S'$
        \EndWhile
    \EndFor
\end{algorithmic}
\end{algorithm}

In optimization algorithms like EAs, Q-learning can guide the search process. The agent's state reflects whether fitness has improved, represented by two states: improvement or no improvement. These states, combined with actions, form a Q-table that evolves during optimization. Actions correspond to selecting search operators, such as crossover and mutation, with Q-values determining operator selection via an $\epsilon$-greedy policy. 

The reward function evaluates the success of actions based on fitness improvements, as shown in Eq.~(\ref{eq:reward}). The Q-values in the Q-table are updated based on the expected maximum reward from the next action, using the Bellman equation shown in Eq.~(\ref{eq:Bellman}).
By dynamically adjusting operator selection, Q-learning enhances exploration and exploitation, improving fitness outcomes and making it a valuable tool for complex problem-solving. 
The algorithm of Q-learning is shown in Algorithm~\ref{alg:q_learning}.
The formula for calculating reward is shown as:
\begin{equation}
    R_t = F_T(S_t, A_t) - F_{t-1}(S_{t-1}, A_{t-1}),
\label{eq:reward}
\end{equation}
where $F_t, F_{t-1}$ are values of fitness at the time $t$ and $t-1$, respectively.
The Bellman equation for updating the Q-values is shown below:
\begin{equation}
    Q_{t+1}(S_t, A_t) = Q_{t}(S_t, A_t) + \alpha (R_t + \gamma max_a Q(S_{t+1}, a) - Q(S_t, A_t)),
\label{eq:Bellman}
\end{equation}
where, $\alpha$ is the learning rate, $\gamma$ is the discount factor.
The detailed Q-learning algorithm is shown in Algorithm~\ref{alg:q_learning}.

%After each generation, the RLGA algorithm generates offspring using Q-learning-guided operators, refining the evolutionary process to achieve optimal solutions.

% Second Algorithm: RLGA
\subsubsection{Reinforcement-enhanced genetic algorithm: RLGA}
GA mimics the process of biological evolution to iteratively search for high-quality solutions through population-based exploration. This population-based search provides strong global exploration capabilities, which is particularly advantageous in solving large-scale, complex optimization problems. However, a key limitation of GA is their sensitivity to parameter configurations. This sensitivity necessitates extensive parameter tuning, making GA highly problem-specific. When the problem domain or scenario changes, parameters must be adjusted or reconfigured, which is often time-consuming. To address these limitations, we propose a reinforcement learning-enhanced genetic algorithm (RLGA) for WFLO. 
In the RLGA, the Q-learning method is used, which plays a critical role in dynamically selecting the genetic operators (e.g., crossover, mutation and mating) for population evolution. The pseudo-code for this RLGA method is shown in Algorithm~\ref{alg:RLGA}. The $F_t$ and $F_{t-1}$ are the current fitness and last fitness values, respectively.

\begin{algorithm}[H]
\caption{RLGA Algorithm}
\label{alg:RLGA}
\small  % Adjusts the font size of the algorithm
\setlength{\abovecaptionskip}{0pt}  % Reduces space above the caption
\begin{algorithmic}[1]
    \State Initialize turbine parameters: \( u_{\inf}, D, C_T \)
    \State Define physics-informed potential layout
    \State Define wake model parameters: \( \theta, a, r, r_1, z_h, z_0, \alpha_{e} \)
    \State Initialize GA: $N_{\text{p}}$, $N$, $P_C$, $M_C$, $C_C$
    \State Initialize Q-learning: \( \alpha, \gamma, \epsilon, Q , A, S\)
    \For{$n_g = 1$ to $N_g$}
        \State Run GA with $A$
        \State $A \gets$ Select action with Q-learning $(\epsilon, s, a)$
        \State $F \gets$ Fitness evaluation $(S_t^p)$ by Eq.~(\ref{eq:fitness})
        \State $R \gets$ Compute reward $(F_t, F_{t-1})$ by Eq.~(\ref{eq:reward})
        \State $S_{t+1} \gets$ Generate the next state using $int( F_t > F_{t-1})$
        \State Select best solutions
        \State Update Q-table using Eq.~(\ref{eq:Bellman}): 
        \State Update state: $S_{t} \gets S_{t+1}$
        \State Update last fitness: $F_{t-1} \gets F_t$
    \EndFor
    % \EndWhile
    \State Post-process and visualize optimized wind farm layout
\end{algorithmic}
\end{algorithm}

\begin{figure}[H]
    \centering
    \includegraphics[width=1\textwidth]{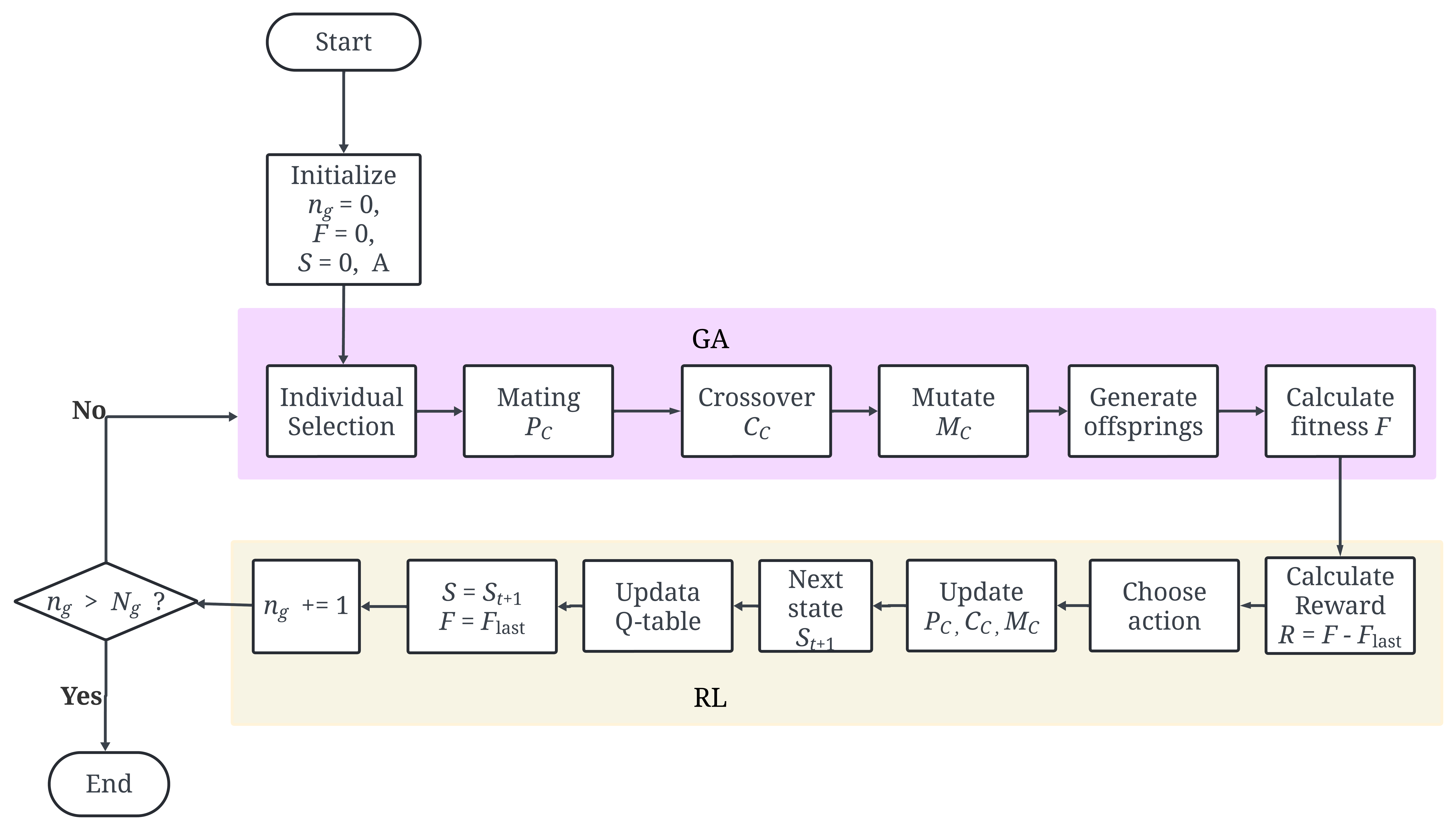}
    \caption{The flow chart of the RLGA method with GA and RL algorithms shown in pink and blue boxes, respectively.}
    \label{fig:flow_chart}
\end{figure}

Fig.~\ref{fig:flow_chart} illustrates the flow chat of the RLGA, showing the data flow between the Q-learning process and the GA process. The upper pink box represents the population evolution process driven by the GA, while the bottom blue box represents the Q-learning agent, which selects the appropriate operators and updates the state based on feedback from the population. As shown, there is a close interplay between GA and Q-learning, where GA provides feedback to the RL agent, and the actions chosen by the agent are applied directly to the GA’s population, ensuring that operator selection is guided by the optimization process itself.

\section{Results and discussion} \label{sec:results}
In this section, the proposed RLGA is validated using a classical WFLO case. To demonstrate the advantages of the RLGA, its convergence performance is compared with that of the GA. Finally, the RLGA is applied to the WFLO of a large wind farm.

\subsection{Case setups}\label{sec:Potential_layout}
\begin{figure}[H]
    \centering
    \includegraphics[width=0.45\textwidth]{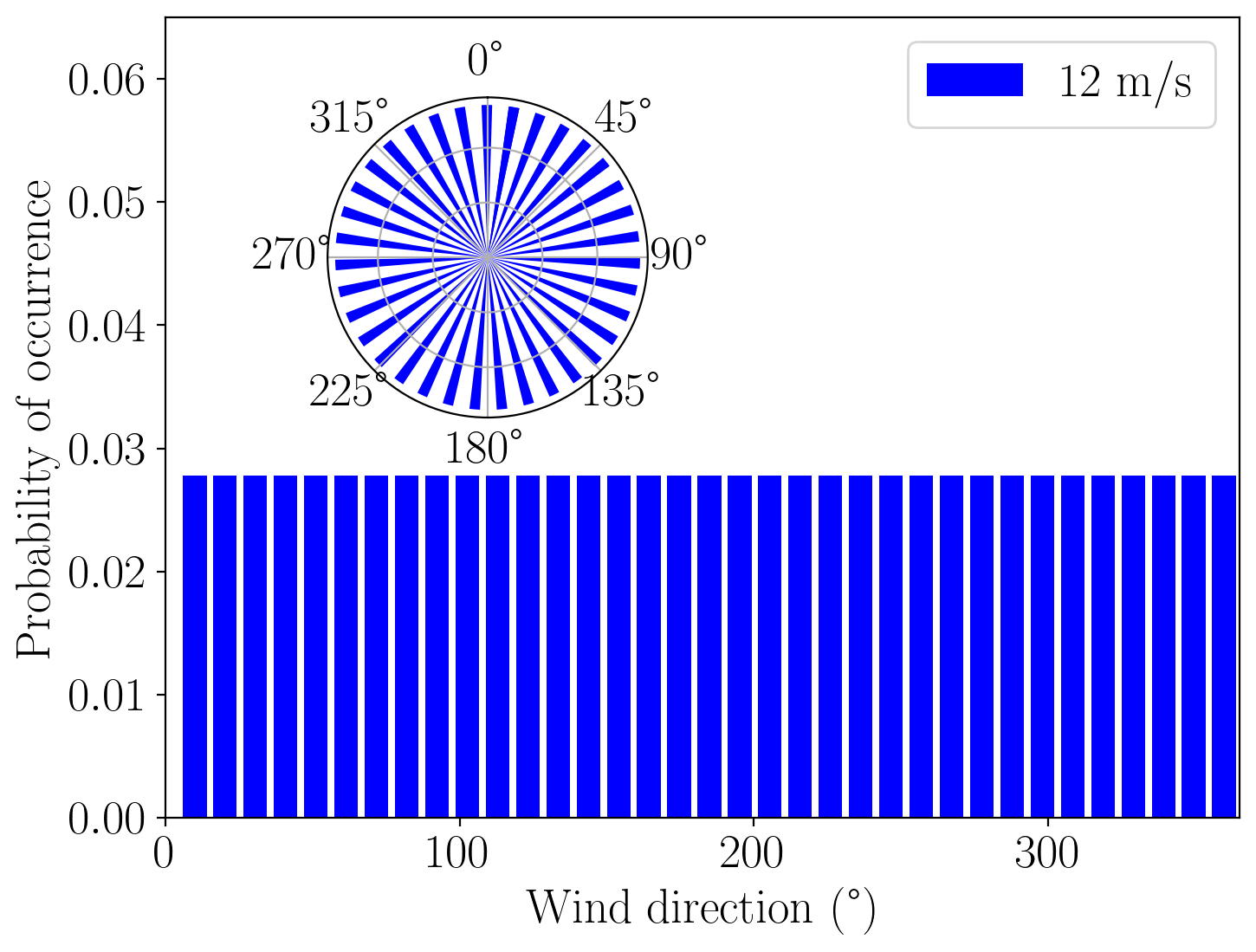}
    \includegraphics[width=0.45\textwidth]{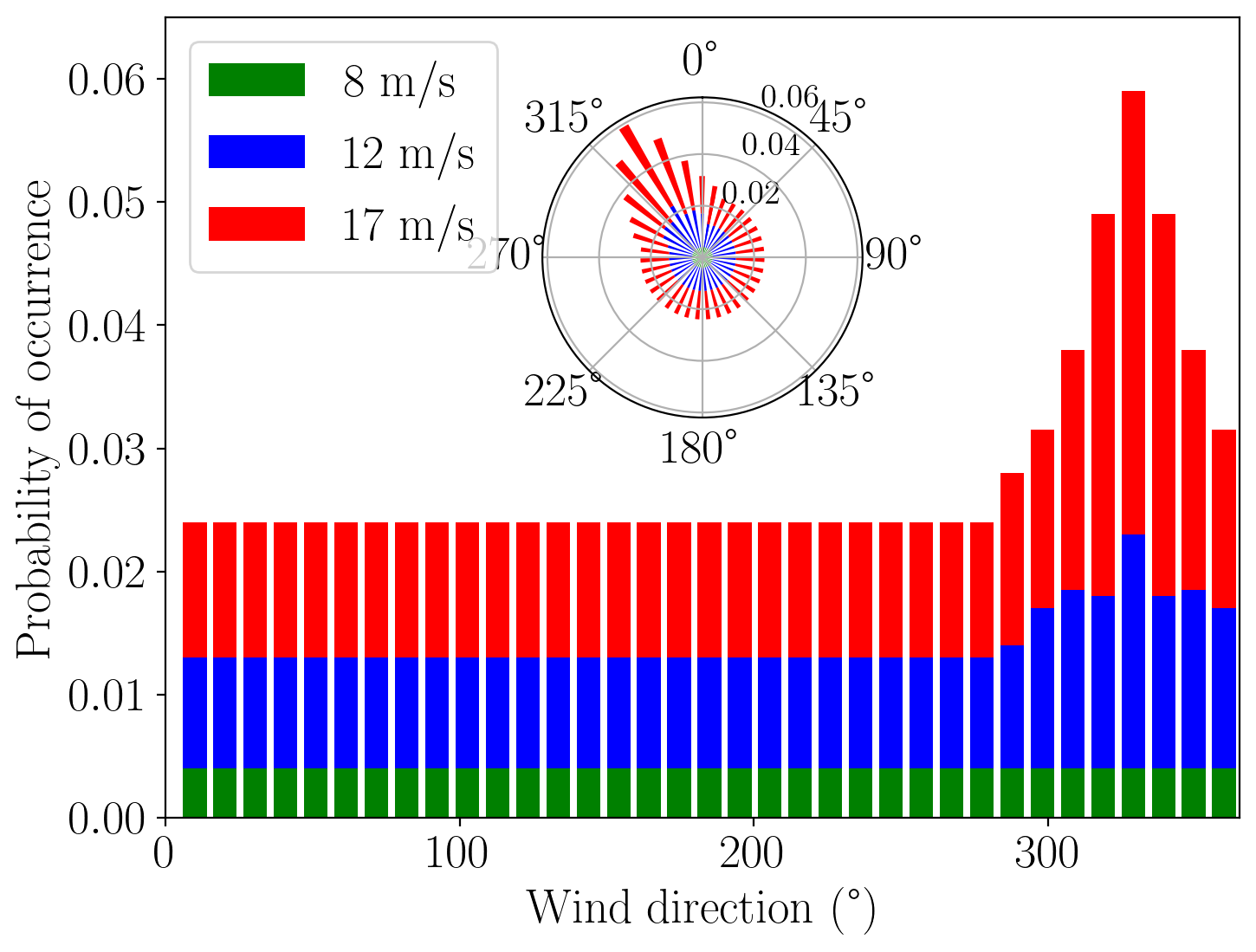}
    \caption{The wind rose for cases the omnidirectional uniform wind (left) and spread non-uniform wind (right).}
    \label{fig:WindRose-CaseBC}
\end{figure}

Three wind scenarios are considered in this work, with the corresponding wind roses shown in Fig.~\ref{fig:WindRose-CaseBC}, where $0^\circ$ represents North~\cite{grady2005placement, wu2021design, pookpunt2013optimal, abdelsalam2018optimization}.
Case A: unidirectional uniform wind with a wind speed of 12 m/s coming from the North. Case B: omnidirectional uniform wind with a mean wind speed of 12 m/s coming from all directions. Each of the 36 angles under consideration represent $10^\circ$ increments from $0\circ$ to $360\circ$ each of which has an equal fraction of occurrence. Case C: spread non-uniform wind as indicated by the wind rose, which is also used by Grady et al.~\cite{grady2005placement} and Parada et al.~\cite{parada2017wind}. As observed, wind speeds of 12 m/s and 17 m/s are predominant, particularly between the angles of 280$^{\circ}$ and 360$^{\circ}$.
Similar to our previous work~\cite{wu2021design}, four physics-informed layouts (aligned, staggered, sunflower, and unstructured) are also considered.
For the wind A scenario, a staggered arrangement can be easily achieved using a simple staggered mesh. In this mesh, potential turbine positions are defined by shifting every other row to the right by half the grid width. However, when the wind blows from all directions, a staggered grid becomes less straightforward. Therefore, the sunflower and unstructured meshes are explored. The details of how these meshes are generated can be found in Ref.~\cite{wu2021design}.
The staggered mesh approach perfectly aligns turbine locations for north-south wind directions, while the unstructured and sunflower mesh approaches aim to provide staggered arrangements for winds blowing from various directions.

The proposed RLGA method is validated by comparing it with our previous work~\cite{wu2021design}, where approximately 100 potential turbine positions are considered in a 2 km $\times$ 2 km wind farm with $\Delta x = \Delta y = 5D$ in the aligned physics-informed layout. To demonstrate the advantages of the RLGA in dynamically selecting GA parameters for rapid convergence, two more complex cases are considered, resulting in approximately 625 and 900 potential turbine positions. Case I: $\Delta x = \Delta y = 2D$ in the 2 km $\times$ 2 km wind farm; Case II: $\Delta x = \Delta y = 5D$ in the same wind farm. A summary of all cases is provided in Table~\ref{tab:All_cases}.

In the uniform wind and non-uniform wind conditions, an ideal value for the objective function, $f_{\text{obj,ideal}} = 1.286 \times 10^{-3} / \text{kW}$ and $6.785 \times 10^{-4} / \text{kW}$, can be be can be determining as:
\begin{equation}
    f_\text{obj, ideal} = \frac{1}{0.3 \sum_i^n u_i^3} \left( \frac{2}{3} + \frac{1}{3} \exp \left(-0.00174N^2\right) \right)
    \label{eq:obj}
\end{equation}
for $N \rightarrow{\infty}$, $u_i$ can be identified in the wind roses shown in Fig.~\ref{fig:WindRose-CaseBC}

\begin{table}
    \centering
    \caption{All cases studied in the present work.}
    \begin{tabular}{c|c|c|c} \hline \hline
Wind farm size & case & Wind Directions & $\Delta x$ \\ \hline
\multirow{3}{6em} {2 km $\times$ 2 km} & Case IA & Unidirectional uniform  
    & $5D$ (Coarse layouts)   \\ 
                                     & Case IB & Omnidirectionaluniform 
    & $5D$ (Coarse layouts)   \\  
                                     & Case IC & Spread non-uniform
    & $5D$ (Coarse layouts)   \\ \hline
\multirow{3}{6em} {2 km $\times$ 2 km} & Case IIA & Unidirectional uniform  
    & $2D$ (Fine layouts)   \\ 
                                     & Case IIB & Omnidirectionaluniform 
    & $2D$ (Fine layouts)   \\  
                                     & Case IIC & Spread non-uniform
    & $2D$ (Fine layouts)   \\ \hline
\multirow{3}{6em} {6 km $\times$ 6 km} & Case IIIA & Unidirectional uniform  
    & $5D$ (Coarse layouts)  \\ 
                                     & Case IIIB & Omnidirectionaluniform 
    & $5D$ (Coarse layouts)  \\  
                                     & Case IIIC & Spread non-uniform
    & $5D$ (Coarse layouts)   \\ \hline \hline
      \end{tabular}
    \label{tab:All_cases}
\end{table}

% \begin{table}
%     \centering
%     \caption{The physics-informed potential layouts. }
%     \begin{tabular}{c|c|c} \hline \hline
%     &  Layouts & No. of potential points  \\ \hline
% \multirow{4}{4em}{Case I} & Aligned      & 100  \\ 
%         & Staggered    & 100 \\  
%         & Sunflower    & 103 \\
%         & Unstructured & 103 \\ \hline
% \multirow{4}{4em}{Case II}   & Aligned        & 625 \\
%         & Staggered    & 625 \\
%         & Sunflower    & 627 \\
%         & Unstructured & 626 \\  \hline
% \multirow{4}{4em}{Case III}   & Aligned        & 900 \\
%         & Staggered    & 900 \\
%         & Sunflower    & 901 \\
%         & Unstructured & 901 \\  \hline         \hline
%       \end{tabular}
%     \label{tab:All_potentail_layouts}
% \end{table}

\subsection{Validation} \label{sec:validation}
To validate the implementation of the current RLGA algorithm for WFLO problems, the results from the aligned case computed in this study are compared with those from Wu~et~al.~\cite{wu2021design}. The quantitatively results of the optimal layouts obtained from different cases are compared in Table \ref{tab: Case_IA_Results}. 

\begin{table}[H]
    \centering
    \caption{Results obtained in the present work and in Wu~et~al.~\cite{wu2021design} for WFLO under unidirectional uniform wind in a small wind farm. }
    \begin{tabular}{c|c|c|c|c} \hline \hline
                  &  Aligned & Staggered & Sunflower &Unstructured  \\ \hline
      $P_{total}$(kW)  & 14310 & 19898 & 20631 & 20096 \\ 
      $P_{total}$(kW)~\cite{wu2021design}.  & 14310 & 19898 & 20551 & 18605 \\ \hline
      $f_\text{obj}$ & 0.0015436 & 0.0013816 & 0.0013887 & 0.0013966 \\
      $f_\text{obj}$~\cite{wu2021design}) & 0.0015436 & 0.0013816 & 0.0013941 & 0.001447 \\ \hline
      $\eta $                    & 0.9202 & 0.9596 & 0.9476 & 0.9455  \\
      $\eta $~\cite{wu2021design} & 0.9202 & 0.9596 & 0.9439 & 0.9203  \\ \hline
      $N$ & 30 & 40 & 42 & 41 \\ 
      $N$~\cite{wu2021design}& 30 & 40 & 42 & 39 \\ \hline \hline
      \end{tabular}
    \label{tab: Case_IA_Results}
\end{table}

As observed, the optimal solutions from the aligned and staggered cases are identical to those obtained by Wu et al.~\cite{wu2021design}, validating our results. Interestingly, for the sunflower and unstructured layouts, the $f_\text{obj}$ value obtained in this work is also lower than that reported in the recent study by Wu et al.~\cite{wu2021design}, with approximately 0.4\% and 8\% improvements in power output for the sunflower and unstructured layouts, respectively. 

\begin{figure}[H]
    \centering
    \includegraphics[width=1\textwidth]{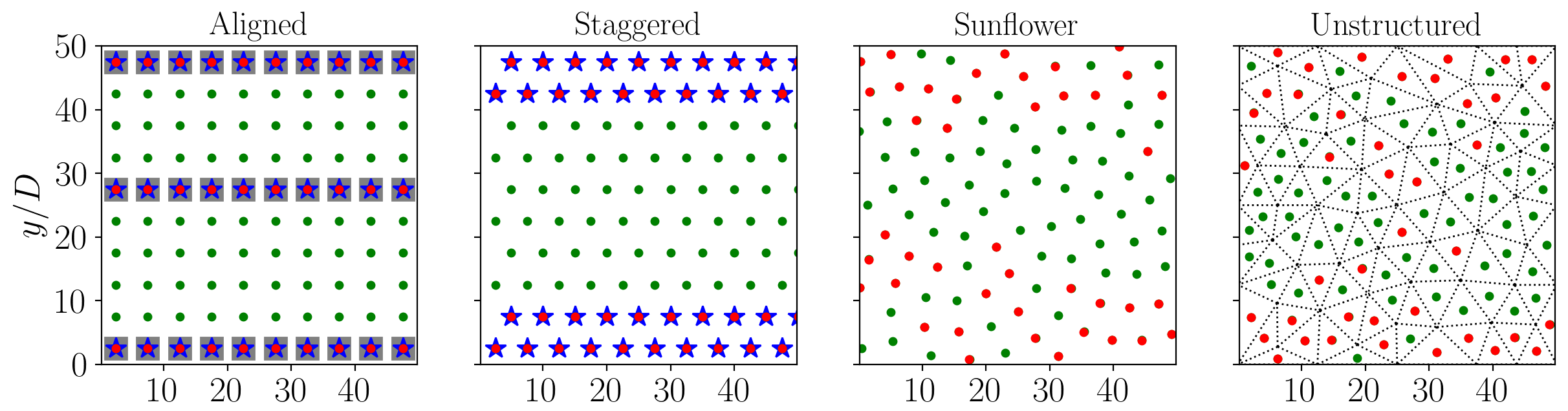}
    \includegraphics[width=1\textwidth]{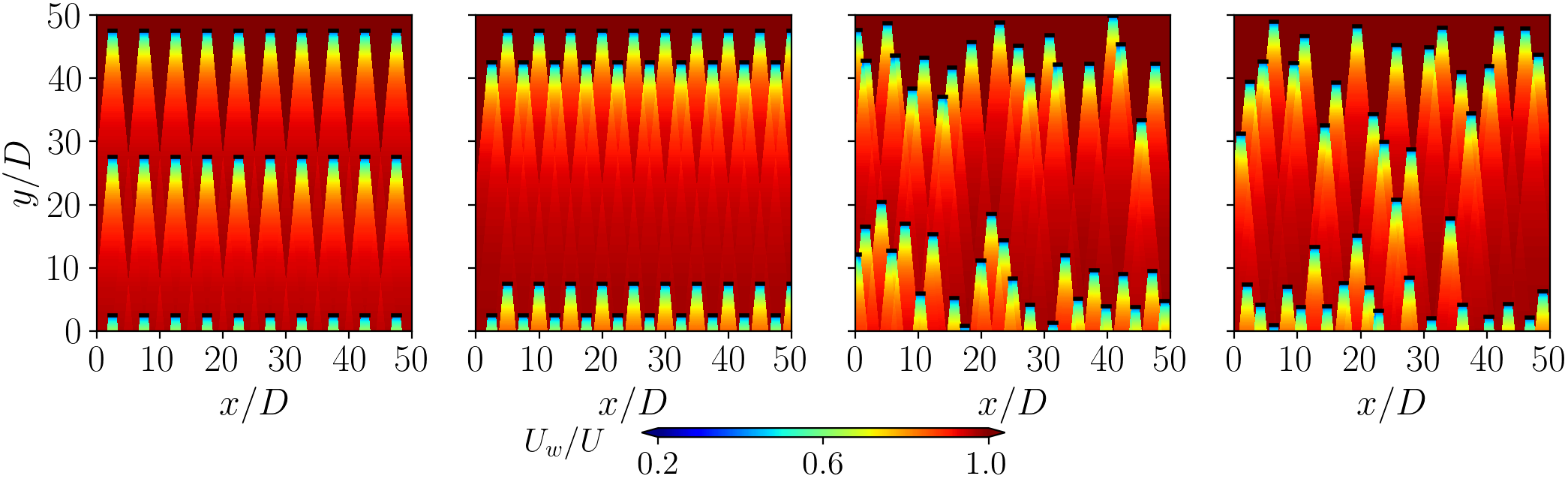}
    \caption{Optimal layouts (the first row) and the corresponding contours (the second row) obtained from the RLGA algorithm in Case IA. Green dots: potential positions; Gray squares: Grady et al.~\cite{grady2005placement}; Blue stars: Wu et~al.~\cite{wu2021design}; Red dots: present.}
    \label{fig:Case_IA_Layouts}
\end{figure}

Fig.~\ref{fig:Case_IA_Layouts} compares the layouts obtained in the present work and in Refs.~\cite{grady2005placement, wu2021design} and shows the corresponding contours when unidirectional uniform wind blowing from north. As seen, the downwind velocity contours for the staggered, unstructured, and sunflower mesh layouts are more complex compared to the aligned mesh layout, where fewer turbines are positioned in the center of the domain.
In the aligned configuration, the results obtained in this work are consistent with those reported by Grady~et~al.~\cite{grady2005placement} and Wu~et~al.~\cite{wu2021design}. Similarly, for the staggered configuration, our findings align with those of Wu~et~al.~\cite{wu2021design}, further validating the accuracy of our results.

\subsection{Comparison of convergence efficiency between RLGA and GA.}\label{sec:compare}
To quantitatively show the efficiency of the proposed RLGA in dealing with the WFLO, the convergence of the RLGA and GA are compared in this section. 
\begin{table}[H]
    \centering
    \caption{Parameter used for GA and RLGA.}
    \begin{tabular}{c|c|c} \hline \hline
                  &  GA & RLGA  \\ \hline
      $N_{p}$ &   5 &   5    \\ 
      $P_c$       &   2 &   [2,3] \\
      $C_c$       & $``\text{single point}"$ & [$``\text{single point}"$, $``\text{uniform}"$, $``\text{two points}"$, $``\text{scattered}"$] \\
      $M_c$       &   4 & [1,2,3,4] \\ 
\hline \hline      
      \end{tabular}
    \label{tab:RLGA_GA_Compare}
\end{table}

In the comparison of RLGA and GA, the parameters utilized during the GA and RLGA iterations are summarized in Table~\ref{tab:RLGA_GA_Compare}. In this table, $P_C$ denotes the number of parents mating, $C_C$ specifies the crossover types, and $M_C$ indicates the mutation percentage of genes. These parameters are held constant throughout the GA process; however, during the RLGA process, they are dynamically adjusted based on the provided options and previous results. 
\begin{figure}
    \centering
    \includegraphics[width=1\linewidth]{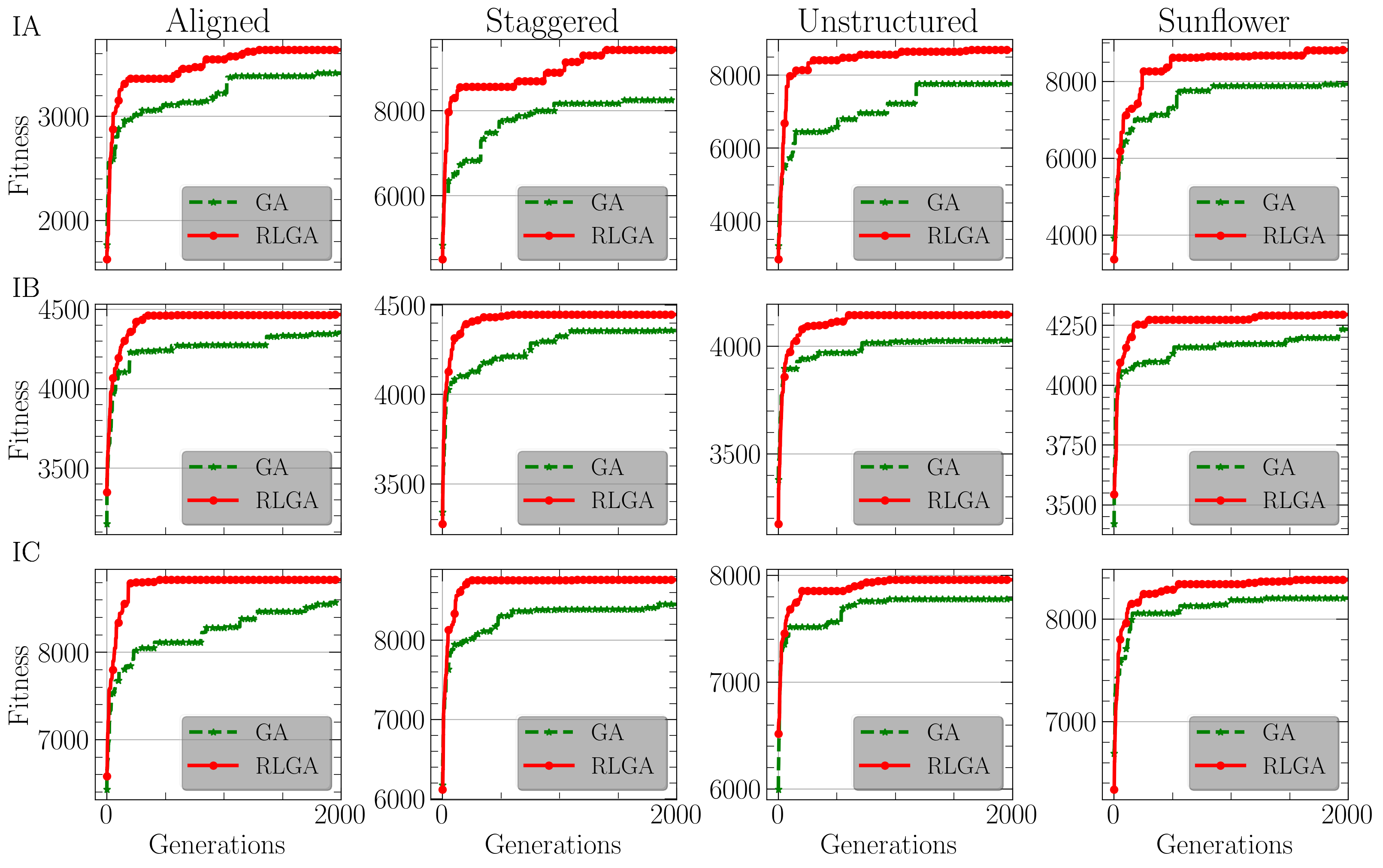}
    \caption{The comparison of convergence efficiency between GA and RLGA in optimization case IA (the first row), IB (the second row) and IC (the last row). }
    \label{fig:Compare_GA-RLGA_I}
\end{figure}

\begin{figure}
    \centering
    \includegraphics[width=1\linewidth]{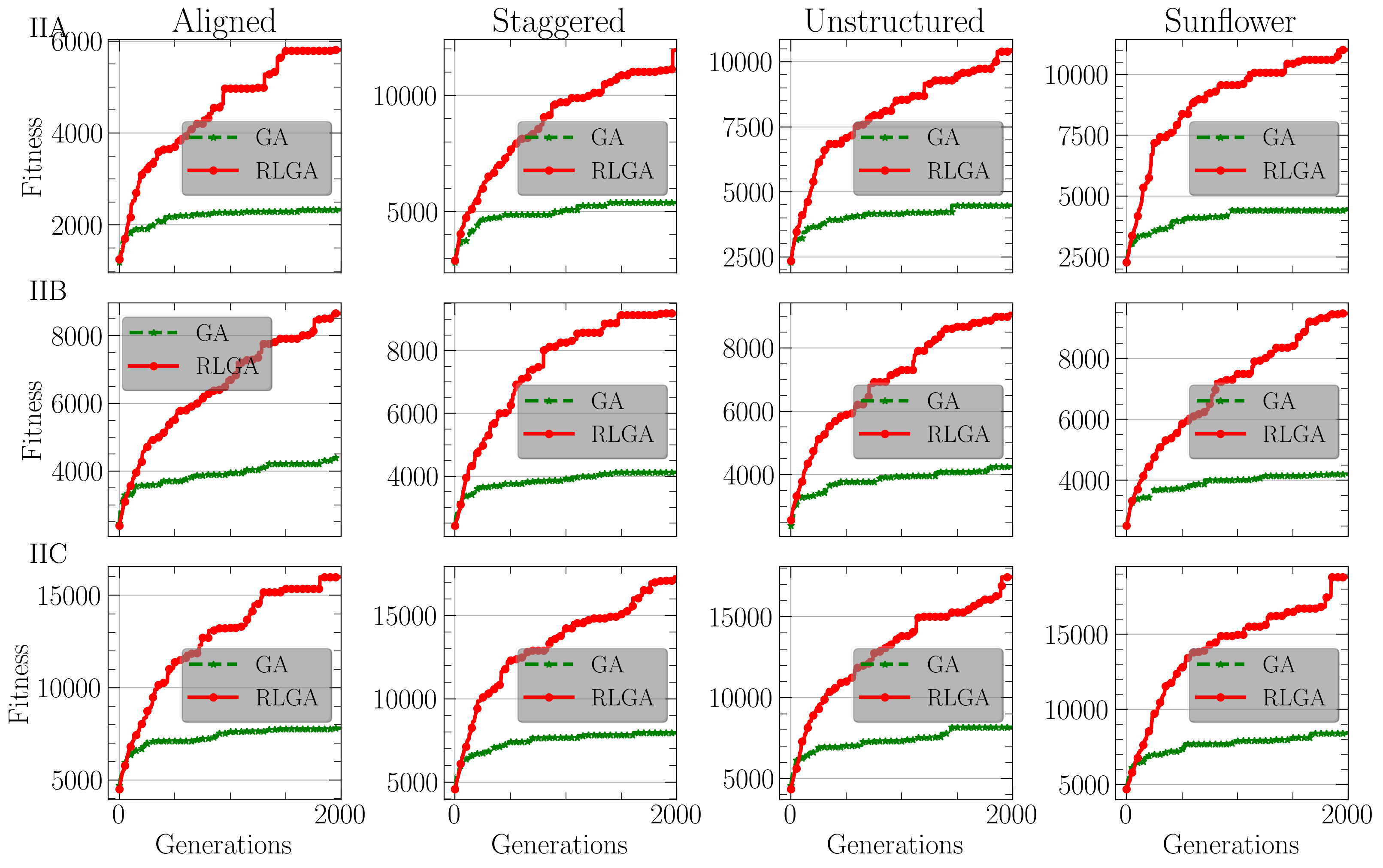}
    \caption{The comparison of convergence efficiency between GA and RLGA in optimization case IIA (the first row), IIB (the second row) and IIC (the last row).}
    \label{fig:Compare_GA-RLGA_II}
\end{figure}

\begin{figure}
    \centering
    \includegraphics[width=1\linewidth]{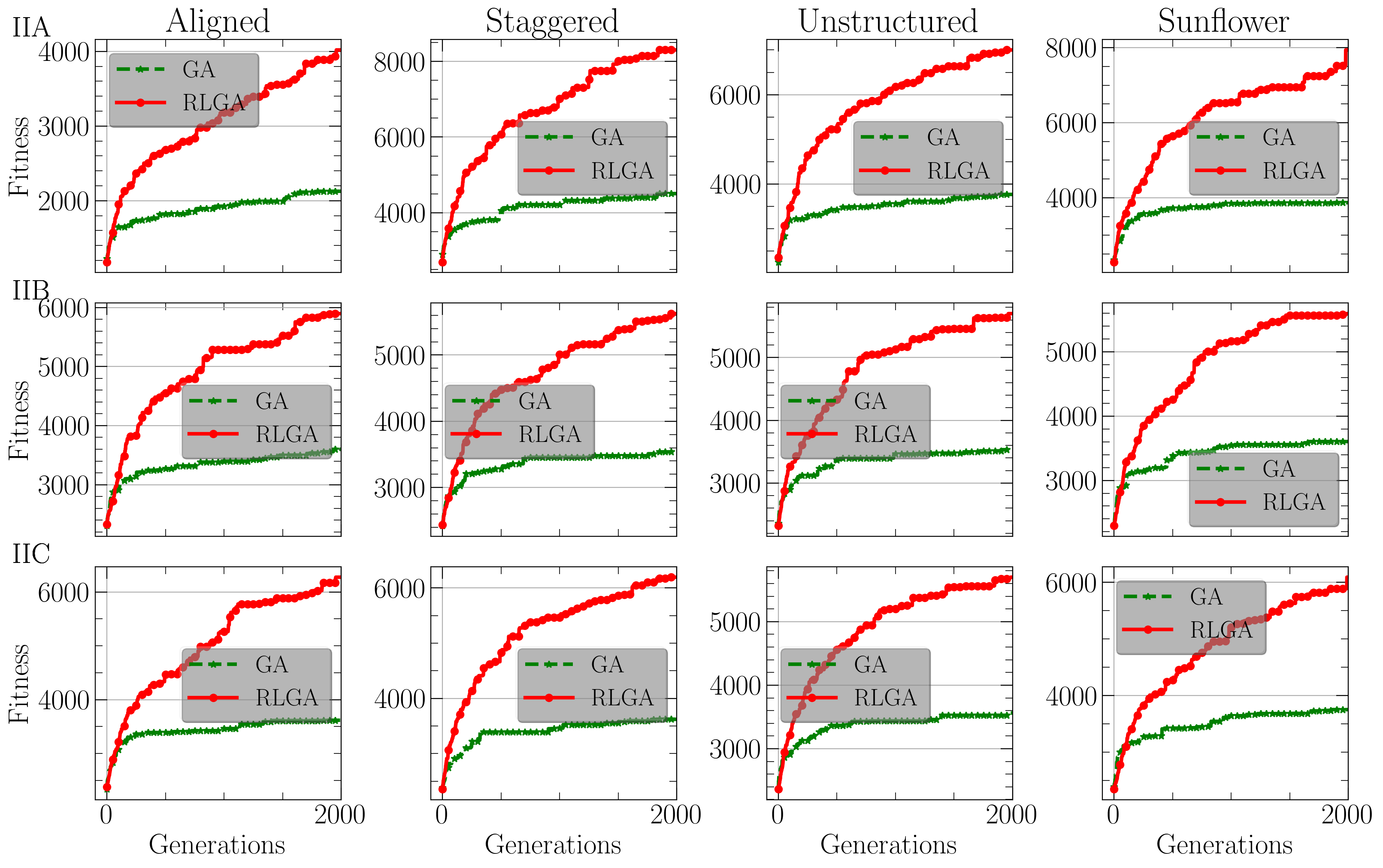}
    \caption{The comparison of convergence efficiency between GA and RLGA in optimization case IIIA (the first row), IIIB (the second row) and IIIC (the last row).}
    \label{fig:Compare_GA-RLGA_III}
\end{figure}

Fig.~\ref{fig:Compare_GA-RLGA_I}, \ref{fig:Compare_GA-RLGA_II}, and \ref{fig:Compare_GA-RLGA_III} illustrate the convergence efficiency of RLGA and GA in addressing WFLO for Cases I, II, and III, respectively. The primary distinction among these cases lies in the number of potential turbines, a critical factor influencing the complexity of the optimization for WFLO. Given the immense complexity of WFLO, even with fewer than 30 turbines, the solution space can exceed $10^{44}$. However, there are approximately 100, 625, and 900 potential turbines for Cases I, II, and III, respectively, indicating a tremendous increase in complexity from Case I to Case III. 
The comparisons indicate that the convergence of RLGA is faster than that of GA, with this advantage becoming more pronounced as the complexity of the optimization increases. In Case I, while the convergence speed of RLGA surpasses that of GA, the benefits of the RLGA method are less significant when addressing smaller problems. However, as complexity grows, the advantages of RLGA increase substantially, reaching approximately three times that of GA. This suggests a strong potential for employing RLGA in optimizing larger wind farms, which aligns with the evolving demands of furture wind energy development.

\subsection{Application in large wind farm}\label{sec:Results_CaseIII}
The previous work investigated the WFLO in a small wind farm (2 km $\times$ 2 km). Given the increasing importance of large wind farms in the future of wind energy, it is essential to explore the WFLO problem within this context. This section examines a large wind farm with dimensions of 6 km $\times$ 6 km (approximately nine times the area of the small wind farm), with potential turbine placements at distances of $5D$ in the aligned configuration. 

%\subsubsection{Unidirectional uniform wind} 
For the simplest unidirectional uniform wind condition, the results obtained by the RLGA are presented in Table~\ref{tab: Case_IIIA_Results}. As shown, both the sunflower and unstructured layouts support a higher number of turbines, resulting in greater power outputs. Among these, the sunflower layout achieves the highest power output, approximately 8.5\% greater than that of the aligned layout, which produces the lowest power. This trend is consistent with the results from the small wind farm presented in Table~\ref{tab: Case_IA_Results}, where the aligned layout contains the fewest turbines. The staggered layout exhibits the highest efficiency, followed by the sunflower layout, while the aligned layout demonstrates the lowest efficiency.

\begin{table}[H]
    \centering
    \caption{Results obtained by RLGA for WFLO under unidirectional uniform wind in a large wind farm.}
    \begin{tabular}{c|c|c|c|c} \hline \hline
                   & Aligned & Staggered  & Sunflower & Unstructured\\ \hline
      $P_{total}$(kW) & 50608  & 53308 & 55536 & 54362 \\  \hline
      $f_\text{obj}$ & 0.0013437 & 0.0013131 & 0.0013205 & 0.0013245 \\ \hline
      $\eta$ & 0.9571 & 0.9794 & 0.9739 & 0.9710 \\ \hline
      N & 102  & 105 & 110 & 108 \\  \hline \hline
      \end{tabular}
    \label{tab: Case_IIIA_Results}
\end{table}
\begin{figure}[H]
    \centering
    \includegraphics[width=1\textwidth]{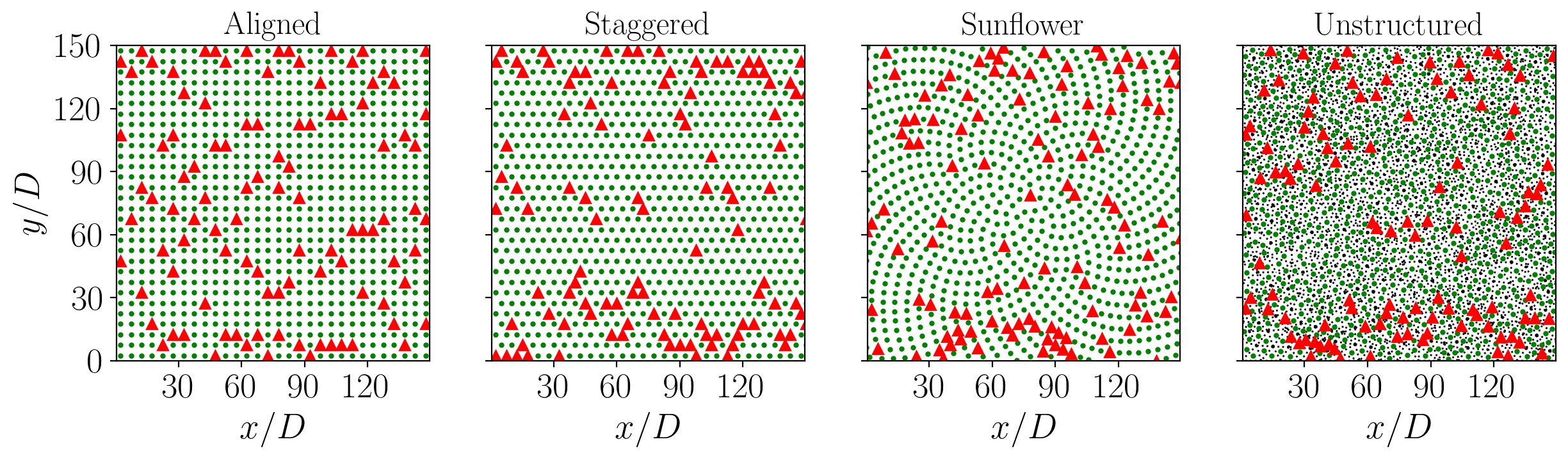}
    \includegraphics[width=1\textwidth]{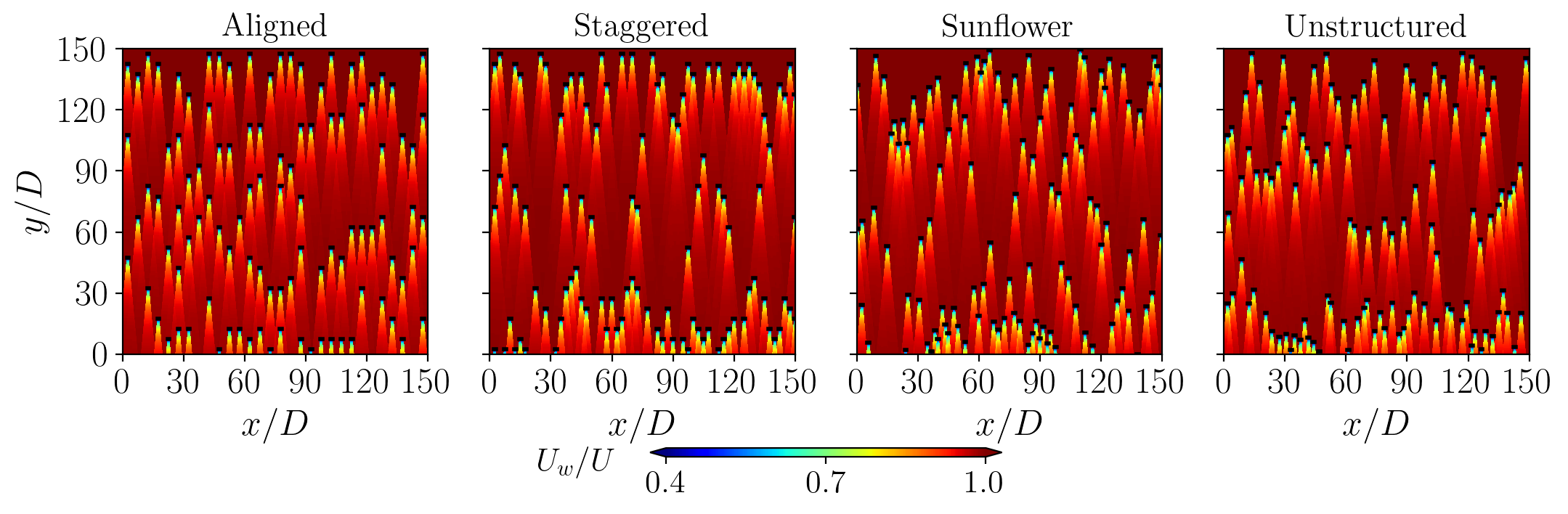}
    \caption{Optimal layouts and contours obtained from the RLGA for WFLO under unidirectional uniform wind in large wind farm. }
    \label{fig:Case_IIIA_Results}
\end{figure} 
The optimal layouts and contours are presented in Fig.~\ref{fig:Case_IIIA_Results}. As observed, turbines are predominantly positioned at the front and rear of the wind farm. The velocity fields are complex across all cases. Compared to results obtained in the small wind farm shown in Fig.~\ref{fig:Case_IA_Layouts}, a greater number of turbines are positioned at the rear of the wind farm in the direction of the wind flow.

%\subsubsection{Omnidirectional uniform wind}
The complexity of real wind environments is significant, as a wind farm site characterized by a prevailing wind direction may still experience winds originating
from various angles. This part examines the phenomenon of omnidirectional uniform wind, which refers to wind uniformly distributed across multiple
directions as depicted in the wind rose shown in Fig.~\ref{fig:WindRose-CaseBC}
Then, the results are presented in Table~\ref{tab: Case_IIIC_Results}.

\begin{table}[H]
    \centering
    \caption{Results obtained by RLGA for WFLO under omnidirectional uniform wind in a large wind farm.}
    \begin{tabular}{c|c|c|c|c} \hline \hline
                   & Aligned & Staggered  & Sunflower & Unstructured\\ \hline
      $P_{total}$(kW) & 41288  & 39856 & 40352 & 39974 \\  \hline
      $f_\text{obj}$ & 0.0013725 & 0.0013716 & 0.0013713 & 0.0013676 \\ \hline
      $\eta$ & 0.9370 & 0.9376 & 0.9378 & 0.9404 \\ \hline
      $N$ & 85  & 82 & 83 & 82 \\  \hline \hline
      \end{tabular}
    \label{tab: Case_IIIB_Results}
\end{table}
\begin{figure}[H]
    \centering
    \includegraphics[width=1\textwidth]{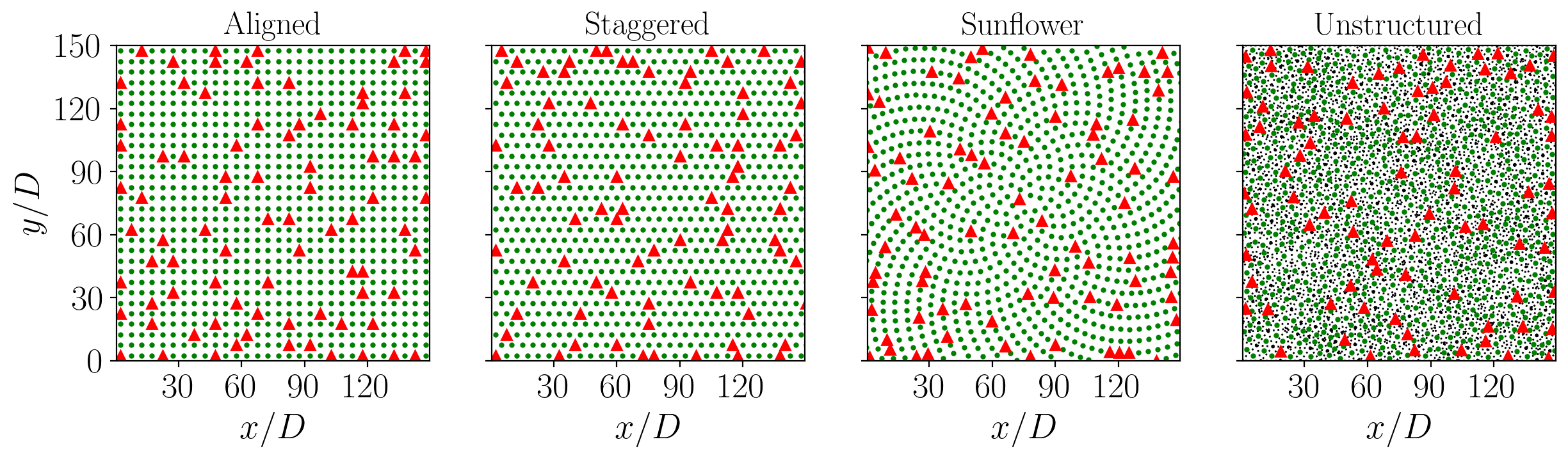}
    \includegraphics[width=1\textwidth]{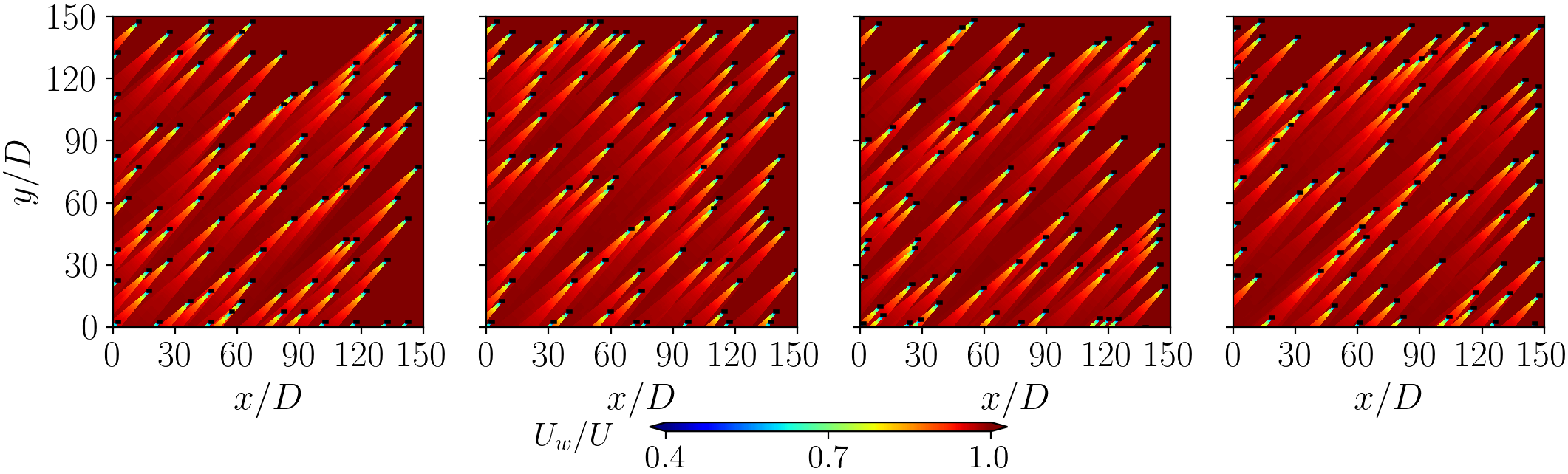}
    \caption{Optimal layouts and contours obtained from the RLGA for WFLO under omnidirectional uniform wind in large wind farm. }
    \label{fig:Case_IIIB_Results}
\end{figure} 

As seen, the aligned layout accommodates the largest number of turbines, resulting in the highest power output, which is approximately 3.6\% greater than that of the staggered layout. Although the number of turbines is identical in the staggered and unstructured layouts, the power output of the unstructured layout is higher, indicating the wake effects are mitigated in the unstructured layout.
The objective values for the sunflower and unstructured layouts are lower than those for the aligned and staggered layouts, while their efficiency is higher, highlighting the importance of physics-informed layouts.
The optimal layouts and corresponding contours are illustrated in Fig.~\ref{fig:Case_IIIB_Results}. As shown, turbines are predominantly positioned near the boundaries of the wind farm.

\begin{table}[H]
    \centering
    \caption{Results obtained by RLGA for WFLO under spread non-uniform wind in a large wind farm.}
    \begin{tabular}{c|c|c|c|c} \hline \hline
                   & Aligned & Staggered  & Sunflower & Unstructured\\ \hline
      $P_{total}$(kW) & 72379  & 72616 & 75983 & 74766 \\  \hline
      $f_\text{obj}$ & 0.0007645 & 0.007620 & 0.0007633 & 0.0007668 \\ \hline
      $\eta$ & 0.8875 & 0.8904 & 0.8889 & 0.8848 \\ \hline
      $N$ & 83  & 83 & 87 & 86 \\  \hline \hline
      \end{tabular}
    \label{tab: Case_IIIC_Results}
\end{table}
\begin{figure}[H]
    \centering
    \includegraphics[width=1\textwidth]{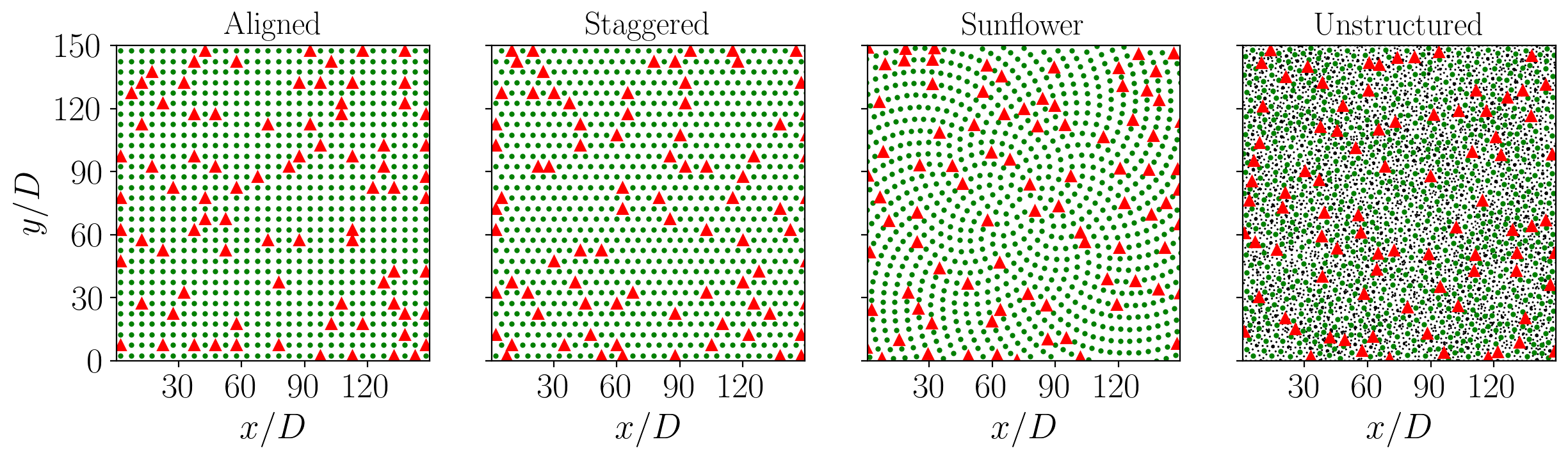}
    \includegraphics[width=1\textwidth]{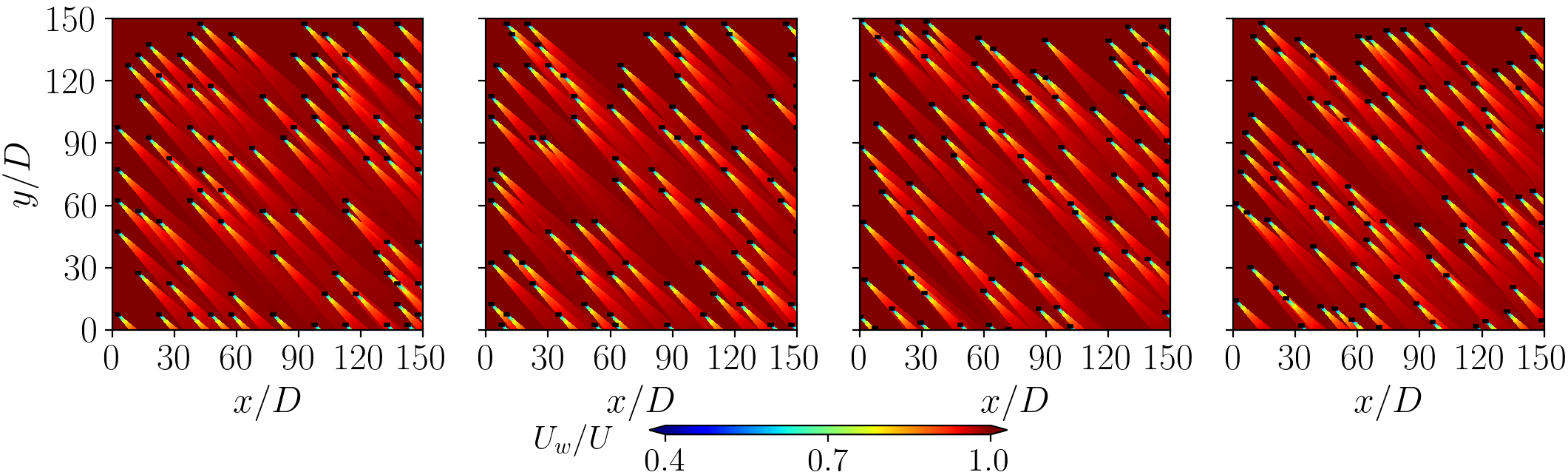}
    \caption{Optimal layouts and contours obtained for WFLO under spread non-uniform wind in large wind farm. }
    \label{fig:Case_IIIC_Results}
\end{figure} 

%\subsubsection{Spread non-uniform wind}
Then, the more complex non-uniform wind are exmained. The result obtained in the spread non-uniform wind are present in Table~\ref{tab: Case_IIIC_Results}.
The sunflower layouts generates the highest power, which is approximately 5.0\% larger than that of the smallest power in the aligned layout. The optimal layouts and corresponding contours are illustrated in Fig.~\ref{fig:Case_IIIC_Results}. As shown, turbines are predominantly positioned near the boundaries of the wind farm, aligned with the prevailing wind conditions ($270^{\circ} - 350^{\circ}$).

\section{Conclusions} \label{sec:conclusion}
Wind farm layout optimization (WFLO) is essential for improving the performance of wind farms. Among the various optimization methods for %employed to tackle 
WFLO, the genetic algorithm (GA) is one of the most widely used techniques~\cite{mosetti1994optimization, grady2005placement, parada2017wind, wu2021design}. However, the optimization efficiency of GA is highly sensitive to the selected parameters, including mating, crossover, and mutation. To improve its efficiency, 
%address this limitation, 
we propose to employ the reinforcement learning (RL), specifically Q-learning %integrated 
to dynamically and automatically select GA parameters based on the performance of previous generations.
%This integration results in the RLGA algorithm, representing the first instance of RL being embedded within GA to address the WFLO problem.

%To evaluate 
The accuracy and robustness of the proposed RLGA algorithm are first evaluated using%, the WFLO problem for 
four physics-informed layouts, i.e., aligned, staggered, unstructured, and sunflower layouts 
under unidirectional uniform wind conditions, which have been employed in our
%is analyzed and compared with 
recent work~\cite{wu2021design}. In the setup, %this study, 
the spacing between neighbor potential turbine locations is $5D$ (where $D$ denotes the turbine diameter) in both streamwise and spanwise directions. 
%in the aligned configuration, set at $5D$ (where $D$ denotes the turbine diameter), i.e., $\Delta x = \Delta y = 5D$. 
The size of the wind farm region %dimensions are specified as 
is $50D \times 50D$ (Case I). %,referred to as  in this paper. 
A good performance is shown for the proposed RLGA algorithm from the comparison with our previous work %are identical to those obtained in Ref.
~\cite{wu2021design}.
%for the aligned and staggered layouts, while outperforming the results for the unstructured and sunflower layout

To illustrate the convergence efficiency and advantages of the proposed RLGA when addressing more complex problems, we investigate two more types of cases, one with a fine mesh spacings ($\Delta x = \Delta y = 2D$, referred to as case II) but in a small wind farm region, the other with a coarse mesh spacing but in a large wind farm region (150$D \times$ 150$D$, referred to as case III), resulting in approximately 625 potential turbines and 900 turbines, respectively. 
%Additionally, 
Three wind conditions are considered: A) unidirectional uniform wind, B) omnidirectional uniform wind, and C) spread non-uniform wind. The comparative analysis of convergence between the proposed RLGA and the classical GA indicates a remarkable improvement of the RLGA algorithm, 
%demonstrates a marked improvement in computational efficiency, 
being roughly three times more effective than the GA. %, especially as the complexity of the problems increases. 
This enhancement is attributed to the RL component, which dynamically adjusts parameters, such as mutation rates, based on the outcomes of genetic iterations, %thereby facilitating escapes from 
to avoid local optima and accelerating convergence. Consequently, this approach enables the rapid optimization of layouts for larger and denser wind farms in practical applications. 
%
%The results obtained across all cases were compared and analyzed. Under unidirectional uniform wind conditions, turbines tend to be positioned at both the forefront and the rear of the wind farms, aligning with the wind direction. In the case of omnidirectional wind, turbine placements are distributed along the four boundaries and the center of the wind farm. Conversely, under spread non-uniform wind conditions, turbines are predominantly located along the boundaries, with a greater alignment towards the prevailing wind direction. Notably, the number of turbines increases for cases II and III in unidirectional uniform wind scenarios.

Because of the improved convergence efficiency, the proposed method has advantageous in tackling %the can be further developed to encompass additional critical aspects and design variables necessary for addressing 
realistic WFLO problems, for instance, considering multiple optimization objectives 
%this study employs a simplified cost model that focuses solely on maximizing power production from the wind farm. Future work can incorporate multiple criteria for evaluation
~\cite{guirguis2017gradient, yu2023reinforcement}, and WFLO in areas 
%Furthermore, this work tests only three ideal wind conditions across various meshes for the design of potential turbine positions. Future investigations should explore optimal turbine positions across different wind roses.
%Additionally, this study considers only square-shaped wind farms. Future research should address wind farms 
with irregular shapes and constraints~\cite{sun2019investigation}, which can be considered in its future development.
%of the method. 
%Larger wind farm is becoming the future of wind turbine development, the present work can be extend to more larger and realistic wind farm applications.
\section*{CRediT authorship contribution statement}
\textbf{Guodan Dong:} Conceptualization, Data curation, Formal analysis, Investigation, Methodology, Software, Validation, Visualization, Writing - original draft. 
\textbf{Jianhua Qin:} Conceptualization, Funding acquisition, Software, Methodology, Writing - review $\&$ editing.
\textbf{Chutian Wu:} Software, Methodology - review.
\textbf{Chang Xu:} Investigation, Methodology, Writing - review $\&$ editing.
\textbf{Xiaolei Yang:} Conceptualization, Funding acquisition, Methodology, Supervision, Writing - review $\&$ editing.

\section*{Declaration of competing interest}
The authors declare that they have no known competing
financial interests or personal relationships that could have appeared to influence the work reported in this paper.

\section*{Acknowledgement}\label{sec:acknowledgement}
%\noindent \textbf{Funding: }
This work was supported by National Natural Science Foundation of China (NO.~12172360, 12202456).
In addition, the authors would like to thank Prof. Wenzhong Shen for his insightful discussions regarding the comparison of optimization methods.

\section*{Data availability}
The data supporting the findings of this study are available from the first and corresponding author upon reasonable request.

\section*{Appendix: the derivation of Jensen wake model} \label{sec:appe_jensen}
According to Fig. \ref{fig:Jensen_model}, the law of conservation of momentum for wind turbines can be expressed as:
\begin{equation}
    \pi r_1^2 U_1 + \pi (r_w^2 - r_1^2) U = \pi r_w^2 U_w.
    \label{eq: Mom_cons}
\end{equation}
Based one $1-D$ momentum theory \cite{hansen2015aerodynamics}, the axial induction factor $a$ can be expressed as: 
$2a = 1 - U_1/U$. And the initial wake radius $r_1$ can be expressed as follows:
\begin{equation}
    r_1 = r \sqrt{ \frac{1-a}{1-2a}  },
\end{equation}
where $r$ is the rotor radius.

Then Eq.~(\ref{eq: Mom_cons}) becomes:
\begin{equation}
    \frac{\Delta U}{U} = \frac{ 2 a}{ (1+\alpha_e x/r_1)^2 },
    \label{eq: Jensen_1}
\end{equation}
where $a$ can also be written as:
\begin{equation}
    a = \frac{ 1-\sqrt{1-C_T} }{2},
    \label{eq: a}
\end{equation}
where $C_T$ is the thrust force coefficient.

Then the Jensen single wake model can be written as:
\begin{equation}
    \frac{\Delta U}{U} = \frac{1-\sqrt{1-C_T}}{(1+\alpha_e X/r_1)^2},
\end{equation}

%\section*{References}
%\bibliographystyle{unsrt}
\small
\bibliography{refs}

\end{document}